\documentclass[12pt]{article}

\def\TITLE{Cloud Diffusion\\Part 1: Theory and Motivation}
\usepackage{hyperref}
\hypersetup{
    pdftitle={\TITLE},  
    pdfauthor={Andrew Randono},
    pdfdisplaydoctitle, 
    colorlinks,
    linkcolor=blue, 
    citecolor=red, 
    urlcolor=blue}

\usepackage[dvipsnames]{xcolor}
\usepackage{xcolor}

\usepackage[
    backend=biber,
    style=alphabetic,
    sorting=ynt]{biblatex}
\usepackage{graphicx}
\usepackage{float}
\usepackage[font=small,labelfont=bf,textfont=it]{caption}
\usepackage{amssymb, amsthm}
\usepackage{bm}
\usepackage{amsfonts}
\usepackage{amsmath}
\usepackage{cases}

\newcommand{\beq}{\begin{equation}}
\newcommand{\eeq}{\end{equation}}
\newcommand{\beqa}{\begin{eqnarray}}
\newcommand{\eeqa}{\end{eqnarray}}

\newcommand{\nn}{\nonumber}
\newcommand{\vep}{\varepsilon}

\newcommand{\G}{\Gamma}

\newcommand{\mc}{\mathcal}

\newcommand{\bb}{\mathbb}

\newcommand{\wt}[1]{\widetilde{#1}}

\begin{document}

{\renewcommand{\thefootnote}{\fnsymbol{footnote}}
\title{\TITLE}
\author{
Andrew Randono\footnote{e-mail: {\tt arandono@thespingroup.org}}\\
The Spin Group Research Institute
}
\date{}
}
\maketitle

\setcounter{footnote}{0}

\begin{abstract}
\noindent
Diffusion models for image generation function by progressively adding noise to an image set and training a model to separate out the signal from the noise. The noise profile used by these models is white noise -- that is, noise based on independent normal distributions at each point whose mean and variance is independent of the scale. By contrast, most natural image sets exhibit a type of scale invariance in their low-order statistical properties characterized by a power-law scaling. Consequently, natural images are closer (in a quantifiable sense) to a different probability distribution that emphasizes large scale correlations and de-emphasizes small scale correlations. These scale invariant noise profiles can be incorporated into diffusion models in place of white noise to form what we will call a ``Cloud Diffusion Model". We argue that these models can lead to faster inference, improved high-frequency details, and greater controllability. In a follow-up paper, we will build and train a Cloud Diffusion Model that uses scale invariance at a fundamental level and compare it to classic, white noise diffusion models.
\end{abstract}

\section{Introduction}
Developed over the last decade, diffusion models are by now firmly established among the most powerful generative models. With a wide range of applications including audio synthesis, protein folding, drug design, materials science, and text generation, diffusion models have arguably had the most spectacular impact in Computer Vision, particularly in image and video generation. 

A seminal paper \cite{sohl2015deep} proposed a physics-inspired model trained to learn how an image is slowly corrupted in a controlled noising procedure using thermodynamic principles, which was developed further into a Score-based Generative Network paradigm \cite{song2019generative}. The trained network can then be reversed, seeding the model with pure noise and generating an image gradually over many small denoising steps. Before diffusion models, the state of the art image generators were typically Generative Adversarial Networks (GANs) or Variational Auto-Encoders (VAEs). While GANs can produce high quality images with reasonable computational efficiency in certain circumstances, they suffer from mode-collapse, limiting the diversity of images a single model can create. On the other hand, VAEs can create a wide range of images, also with reasonable computational efficiency, but the images are often blurry and of lower quality. 

The advent of Denoising Diffusion Probabilistic Models (DDPMs) \cite{ho2020DDPM} cemented the more modern diffusion architecture, and efficiency improvements in the Denoising Diffusion Implicit Models (DDIMs) \cite{song2022DDIM} improved speed of inference. Eventually it was realized that score-based and denoising-diffusion models were equivalent \cite{song2021score, luo2022understandingdiffusionmodelsunified}. The models were improved \cite{dhariwal2021diffusion}, and they eventually out-competed contemporary GANs and VAEs. Diffusion Models tend to be less computationally efficient than GANs or VAEs because of the way they incrementally build images over many time steps. However, processor speeds increased over time and many efficiency tweaks to the original architecture were implemented (e.g. the Latent Diffusion architecture \cite{rombach2022high}), making speed less of a problem, trumped by the improvements in image diversity and quality. By now diffusion models are the preeminent AI models for image and video generation \cite{yang2023diffusion, croitoru2023diffusion, cao2024survey}.

\begin{figure}[htbp]
    \centering
    \includegraphics[width=1.0\linewidth]{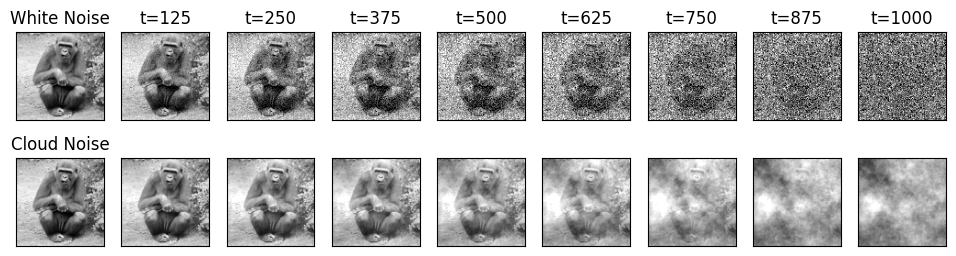}
    \caption{Cloud Diffusion uses scale-invariant noise tuned to the statistical properties of the image set in place of white noise to improve the generative model. Visualized here is the forward diffusion process for white noise diffusion models (top row) and for Cloud Diffusion Models (bottom row).}
    \label{SimpleNoisingProcedure}
\end{figure}

Despite their success, there is still ample room for improvement in the diffusion model paradigm. We propose here that we can do better by using a different type of noise in the noising procedure that is in a certain quantifiable sense `closer' to the image set. Natural images have a peculiar property in their low order statistical properties --- the two-point correlation between neighboring points exhibits a power-law scaling yielding a symmetry under scale transformations. We can model these two-point correlations using a class of noise distributions that exhibit the same scaling properties. The resultant noise falls into a class of scale-invariant normal distributions. We call the scale-invariant noise that is tuned to the statistical parameters of the image set \textbf{Cloud Noise}. We argue here that replacing white noise with Cloud Noise in a diffusion model, as shown in Figure \ref{SimpleNoisingProcedure}, has certain advantages including improvements in performance, quality, and controllability. The model is inspired in part by early universe cosmology where, shortly after the big-bang, primordial density fluctuations exhibit a similar type of power-law scaling. These scale-invariant perturbations collapsed through gravitation attraction, giving rise to the stars, galaxies, and clusters we see today. 

This work is adapted from a series of Jupyter Notebooks and associated code posted in a public GitHub repository \cite{RandonoCloudDiffusionRepo}. Much of the numerical work and codebase is detailed there, and we encourage readers to view the repository. 

\section{Scale Invariance in Natural Image Sets}
Natural image sets exhibit certain statistical regularities in their low-order correlations between neighboring pixels. The term `natural' here is loosely defined, and generally refers to image sets with a wide diversity of subjects (people, animals, vehicles, buildings, trees, etc.) in a variety of perspectives and environmental settings (indoor, outdoor, close-up, landscape, etc.). In aggregate, these image sets exhibit a type of scale invariance in their two-point correlations when analyzed in Fourier Space. This behavior has been known in the literature since at least the mid-eighties, with the first references appearing in 1987 \cite{burton1987color, field1987relations} inspiring follow-up research \cite{tolhurst1992amplitude, ruderman1993statistics, ruderman1994CompNeurSys, ruderman1994statistics, van1996modelling, dong1995statistics}. The power-law scaling is a remarkably robust property of natural images, and has been explained as the result of natural images being comprised of collages of statistically independent objects that have a power-law distribution of sizes \cite{ruderman1997origins}. 

We will not assume familiarity with the preceding work on scale-invariance in natural image sets here. Instead, we introduce the subject through a hypothetical scenario.

\subsection{A Wager}
Imagine for the moment that you are a professional gambler and you are approached by a shady character with a rather odd proposal for a wager. He presents you with a set of black-and-white natural images (satisfying the loose criteria listed above) and tells you that he has selected five images at random. He gives you full access to the image set, but doesn't give you the images that he has selected. Instead, he has selected two pixels at random from each image. He gives you the {\it coordinates} of each pixel. And he gives you the {\it value} of {\it just one} of these pixels. \textbf{Then he challenges you to guess the value of the unknown pixel for each image and to propose an amount to wager.}
\begin{figure}[htbp]
    \centering
    \includegraphics[width=0.8\linewidth]{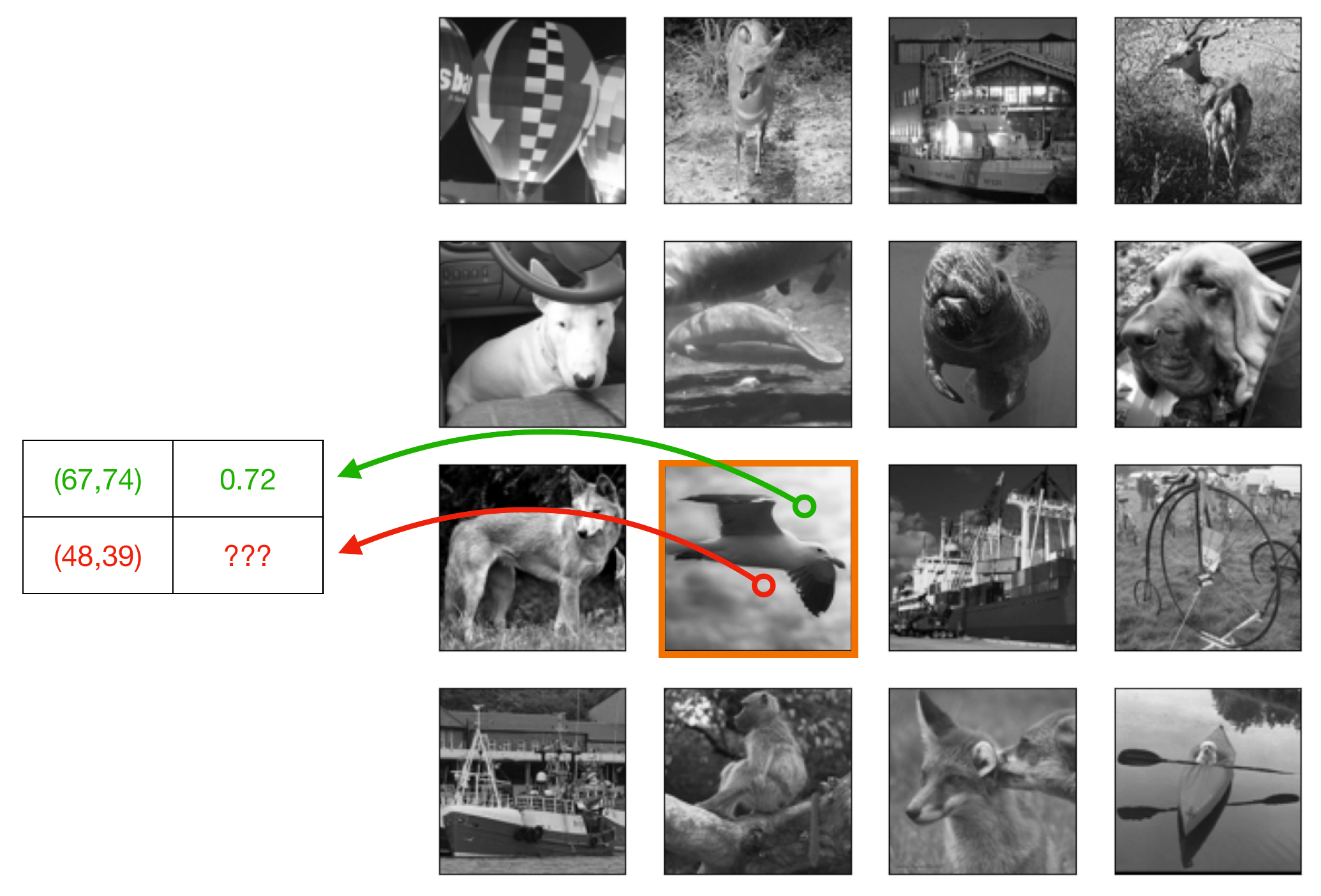}
    \caption{The images are a cleaned subset of the STL10 dataset. A handful of images are selected at random from the image set, and two points from each image are selected at random. You are given the coordinates of the two points, and the value of just one.}
    \label{STL10_Images}
\end{figure}

You are also told the image set is a cleaned subset of the STL10 dataset, with images containing vertical or horizontal bars on the edges (typical artifacts of cropping) removed, resized to $96\times 96$ and converted to grayscale. To summarize, you are given the image set shown in Figure \ref{STL10_Images} score card shown in Table \ref{ScoreCard}.
\begin{table}[h!]
\centering
\begin{tabular}{ |l||c|c||c|c||c|}
\hline
\multicolumn{6}{|c|}{\bf{Score Card}} \\
\hline\hline
& \multicolumn{2}{|c||}{\bf{Pixel \#1}} & \multicolumn{2}{|c||}{\bf{Pixel \#2}} & \\
\hline
 & \bf{(x,y)} & \bf{Value} & \bf{(x,y)} & \bf{Value} & \bf{Wager} \\
\hline
\bf{Image 1} & (12,45) & 0.87 & (82,35) & \bf{\color{red}???} & \bf{\color{ForestGreen}\$\$\$} \\
\bf{Image 2} & (76,12) & 0.25 & (36,48) & \bf{\color{red}???} & \bf{\color{ForestGreen}\$\$\$} \\
\bf{Image 3} & (43,26) & 0.34 & (87,21) & \bf{\color{red}???} & \bf{\color{ForestGreen}\$\$\$} \\
\bf{Image 4} & (87,95) & 0.94 & (89,92) & \bf{\color{red}???} & \bf{\color{ForestGreen}\$\$\$} \\
\bf{Image 5} & (23,41) & 0.56 & (27,19) & \bf{\color{red}???} & \bf{\color{ForestGreen}\$\$\$} \\
\hline
\end{tabular}
\caption{Score card for the hypothetical wager.}
\label{ScoreCard}
\end{table}

\subsection{First Strategy}
\begin{figure}[htbp]
    \centering
    \includegraphics[width=0.8\linewidth]{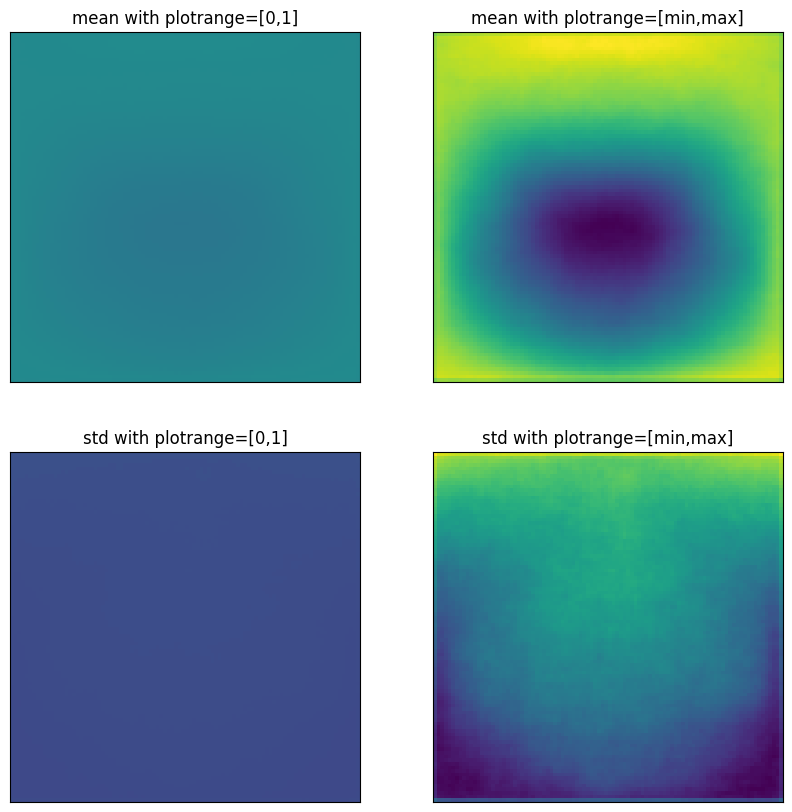}
    \caption{The pixel-wise mean (first row) and standard deviation (second row) of the image set. The first column has plot range [0,1], and the second column is enhanced with plot range from the minimum to the maximum values. The plots reveal that while there is some variation in the mean and standard deviation, they are both extremely uniform over all pixels.}
    \label{fig:MeanAndSTD}
\end{figure}

\begin{figure}[htbp]
    \centering
    \includegraphics[width=0.9\linewidth]{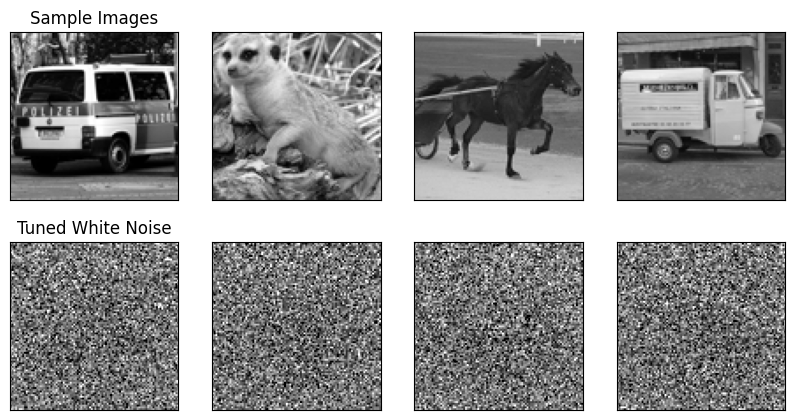}
    \caption{Random samples from the image set shown alongside samples of white noise. Though the noise samples have a pixel-wise mean and standard deviation that is tuned to match that of the dataset, the noise samples capture very little information about the set and look nothing like the image samples.}
    \label{WhiteNoiseSamples}
\end{figure}

On the surface there does not appear to be much information to go off of. But, you at least have the coordinates for Pixel \#2, so you may be able to use that. So your first strategy might be to compute the mean and standard deviation, and try to use this information along with the coordinates of Pixel \#2 to guess its value and choose a dollar amount to wager. With the goal of using the spatial information to inform your guess, you compute the {\it point-wise} mean and the {\it point-wise} standard deviation
\beqa
\mu^{ij} &=& \bb{E}\left[x^{ij}\right] \label{eq:mean}\\
\sigma^{ij} &=& \sqrt{\bb{E}\left[\left(x^{ij}-\mu^{ij}\right)^2\right]} \label{eq:std}
\eeqa
and you visualize these as images as shown in Figure \ref{fig:MeanAndSTD}.
You find that both the mean and standard deviation are exceptionally uniform across all pixels. While there is some bias toward light pixels near the top of the images (to be expected from a natural image set where lighting often comes from above), the effect is very subtle. This gives you some information. You could potentially use this to make a blanket guess about the value of the pixel from the mean, and use the standard deviation to choose the amount to wager. Since the mean and standard deviation are very uniform across all pixels, they can be approximated by
\beq
\mu^{ij} \approx \mu \quad \quad \sigma^{ij}\approx \sigma \quad \quad \text{for all }ij
\eeq
where $\mu$ and $\sigma$ are constants. Correspondingly, your guesses for all pixels should be very similar regardless of the position. You could simply choose the mean at each point for your guess. But a slightly more sophisticated approach would be to model the information you've ascertained as a normal distribution. To do this, you construct a noise generator and model the noise by sampling each pixel from a normal distribution with mean $\mu$ and standard deviation $\sigma$ so that a typical sample is represented by
\beq
x^{ij} \sim \mc{N}(\mu, \sigma^2) \quad \quad \text{for all }ij\,.
\eeq
This distribution looks like pure noise. More specifically, it is a particular type of noise, called \textbf{white noise}. It looks like the images in Figure \ref{WhiteNoiseSamples}. 
This noise looks nothing like a typical image in the data set. Nevertheless, it captures {\it some} of the statistical properties of the image set, namely the mean and standard deviation at each point. So, using this noise distribution as a model, you might try to craft a strategy for guessing the pixel values and make a wager. 

But, this is likely a losing strategy. You haven't used all of the information that you've been given. Can you craft a better strategy that uses the coordinates and value of Pixel \#1 to your advantage? 

\subsection{Second Strategy}
If there were predictable correlations between neighboring points in an image sampled from the data set, then you could possibly use this information to make a guess for the value of Pixel \#2 given its position relative to Pixel \#1. Are there correlations? Likely so. Images of natural scenes and objects have big blocks of color and value. Thus, pixels close to each other probably have similar values. For example, Pixel \#1 of Image 4 is a very bright pixel in the upper right corner of the image. Perhaps the pixel is sampled from the sky in the image? If so, Pixel \#2, which is just a handful of pixels away, is likely to be part of the sky too. So it very well may be a bright pixel, too. How can you model the correlations between pixels? 

The \textbf{Covariance Tensor} can be interpreted as a two-point correlation function --- it is a statistical measure of the correlations between two points in a distribution. For a distribution $\mc{Q}$ it is given by
\beq
{\Sigma^{ij}}_{kl} = \bb{E}_{x\sim \mc{Q}}\left[\left(x^{ij}-\mu^{ij}\right)\left(x_{kl}-\mu_{kl}\right)\right]\,.
\eeq
Note that the diagonal component is related to the standard deviation at each point: ${\Sigma^{ij}}_{ij}=(\sigma^{ij})^2$. When the points at pixel coordinates $\{ij\}$ and $\{kl\}$ are consistently above or below the mean at the same time, the covariance ${\Sigma^{ij}}_{kl}$ is positive. Similarly, when one point is consistently above the mean while the other is consistently below the mean, ${\Sigma^{ij}}_{kl}$ is negative. The magnitude is a measure of the degree of correlation between the two points and their difference from the mean.

\begin{figure}[htbp]
    \centering
    \includegraphics[width=0.9\linewidth]{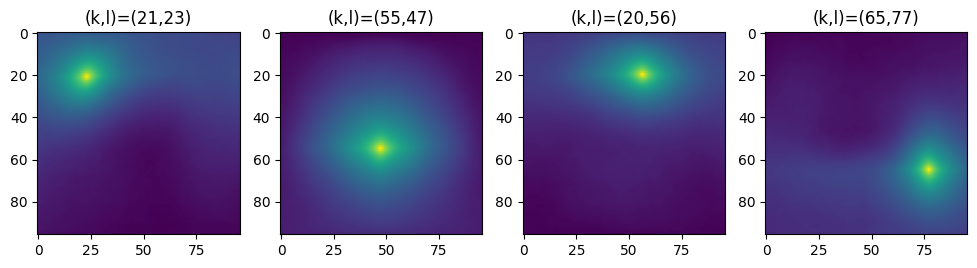}
    \caption{The Covariance Tensor of the image set plotted for four randomly selected points. The plots reveal that neighboring pixels are correlated, and the magnitude of the correlation falls off with radial distance.}
    \label{RealCovariance}
\end{figure}

So, you set about computing the covariance tensor. The result is visualized in Figure \ref{RealCovariance}. In this figure, a handful of coordinates are selected at random for $\{kl\}$ and the remainder is plotted as a 2D image. For example, for the pixel with coordinates $\{kl\}=\{23,74\}$, ${\Sigma^{ij}}_{23\,74}$ is visualized as an image. 
The plots reveal that the correlation between neighboring pixels is highest between immediately adjacent pixels, and it falls off with distance. Moreover, the distance correlation appears to be approximately radial --- it depends only on the magnitude of the distance vector between the points and not on the direction. It also appears that this behavior is approximately uniform --- the radial dependence appears to be almost independent of where the pixels are located in the image. 

With this information, you may be able to craft a better strategy for the bet. For example, you might try to determine the radial dependence of the covariance and and use that to help choose an amount to wager. However, for practical purposes, covariance tensors that are not diagonal (and this one is not) can be difficult to work with. So as an alternative, you might try to diagonalize the covariance tensor to find its eigenvalues, and use the diagonal form to model the image set with noise, similar to how you did it with white noise. Perhaps surprisingly, this task is facilitated by a Fourier Space analysis, which we will do next. 

\subsection{Fourier Space Analysis}
\begin{figure}[htbp]
    \centering
    \includegraphics[width=0.9\linewidth]{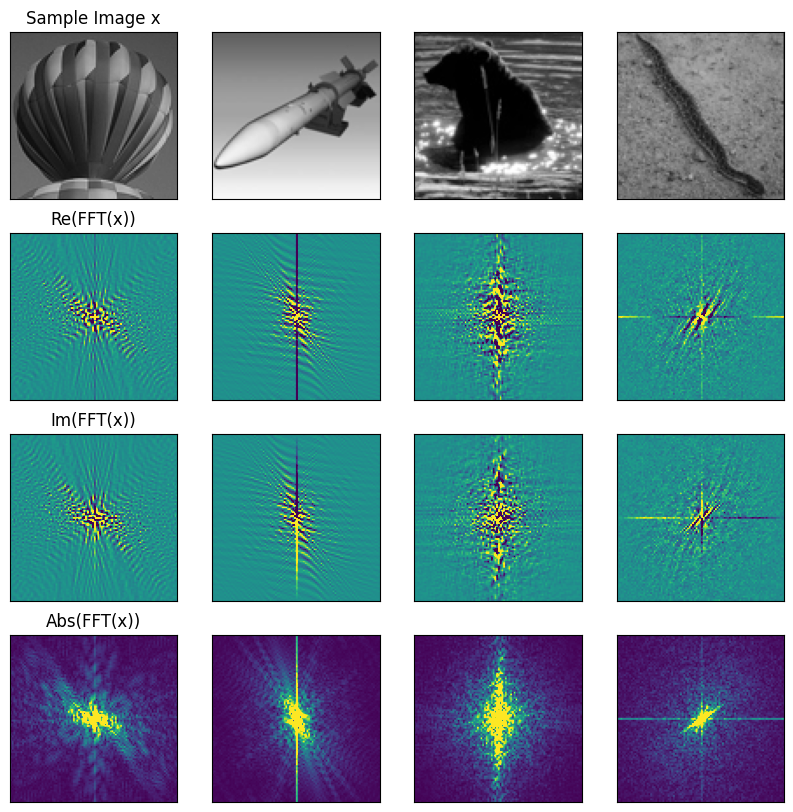}
    \caption{Four random sample images from the dataset with plots of the real part, imaginary part, and absolute value of the Fourier transform of each sample.}
    \label{FourierSamples}
\end{figure}

It so happens that the rich structure of the variance tensor fully emerges when one turns to Fourier Space. In Fourier Space, the covariance tensors (plural as there are two in the complex domain), represent the correlations between modes with different frequencies (or wavelengths/wave-numbers depending on your nomenclature). Samples of the Fourier transformed images are shown in Figure \ref{FourierSamples}. 
A couple of features stick out in these images. In this version of the Fast Fourier Transform, we have moved the zero-wavenumber mode to the center. First, note that the amplitude of the pixels tends to increase towards the center of the images. Second, note the characteristic cross that is prominent in most of the Fourier transformed images. This central cross indicates the presence of features with a horizontal or vertical character. There are three main causes for the central cross:
\begin{enumerate}
\item{Cropping artifacts} \label{item1}
\item{Natural and aritficial horizontal/vertical features}\label{item2}
\item{A-periodic boundary conditions.}\label{item3}
\end{enumerate}
Cropping artifacts usually manifest as letter-boxing --- white or black bars on the top or sides of images. The image set has been cleaned to remove images with prominent letter-boxing, though some letter-boxed images may have slipped through. Natural vertical and horizontal features are present because gravity always points in one direction and it is usually oriented straight down in the images. They include horizon lines, tree trunks, building edges, sign posts, etc. These features can be eliminated from the statistical analysis by applying a random rotation to the images. But this doesn't fix the third and most dominant cause of the central cross. If the image set had periodic boundary conditions, more specifically toroidally periodic boundary conditions where the top/bottom edges match and the right/left edges match, the central cross would be mostly eliminated (barring the effects of the first and second items on the list). Thus, one way to reduce the central cross is to apply a Hamming window --- multiply by an attenuation function that is close to one in the center of the image and smoothly falls off to zero at the edges. Though this is effective, it may alter the overall statistics of the images, which is what we are trying to analyze here. So instead we resort to the more crude method of treating the statistics in the pixels around the horizontal and vertical edges more as outliers, and masking out these values in our analysis.

Finally, we note that although the Fourier transform is complex-valued, since the original images are real, there are relations between the pixels in Fourier space. This effectively reduces the number of complex degrees of freedom by half so that they match the number of real degrees of freedom in the original image. Inverting the image about the horizontal and vertical axes (or equivalently rotating about the center by $180^\circ$) is equivalent to taking the complex conjugate. Letting the tilded coordinates $\{\wt{i}\wt{j}\}$ represent the coordinates $\{ij\}$ rotated about the origin by $180^\circ$, this relation is given by:
\beqa
Z^{ij} &\equiv & \mathcal{F}(x^{ij}) \\
Z^{\wt{i}\wt{j}} &=& (Z^{ij})^*\,.
\eeqa
In Fourier space we can compute the covariance relative to the Fourier-transformed mean, $M^{ij}\equiv \mathcal{F}(\mu^{ij})$. But since the function is complex-valued, there are actually two covariance tensors:
\beqa
{\Gamma^{ij}}_{kl} &=& \bb{E}\left[\left(Z^{ij}-M^{ij}\right)\left(Z_{kl}-M_{kl}\right)^*\right] \\
{C^{ij}}_{kl} &=& \bb{E}\left[\left(Z^{ij}-M^{ij}\right)\left(Z_{kl}-M_{kl}\right)\right]\,.
\eeqa
The two only differ by the the complex-conjugation of the second term, which has the consequence that for real original images, they are related by a $180^\circ$ rotation of one set of indices:
\beq
{C^{ij}}_{kl} = {\Gamma^{ij}}_{\wt{k}\wt{l}}\,. \label{GammaCRelation}
\eeq
So, they are not independent and we can just focus on one of them.

\begin{figure}[htbp]
    \centering
    \includegraphics[width=0.8\linewidth]{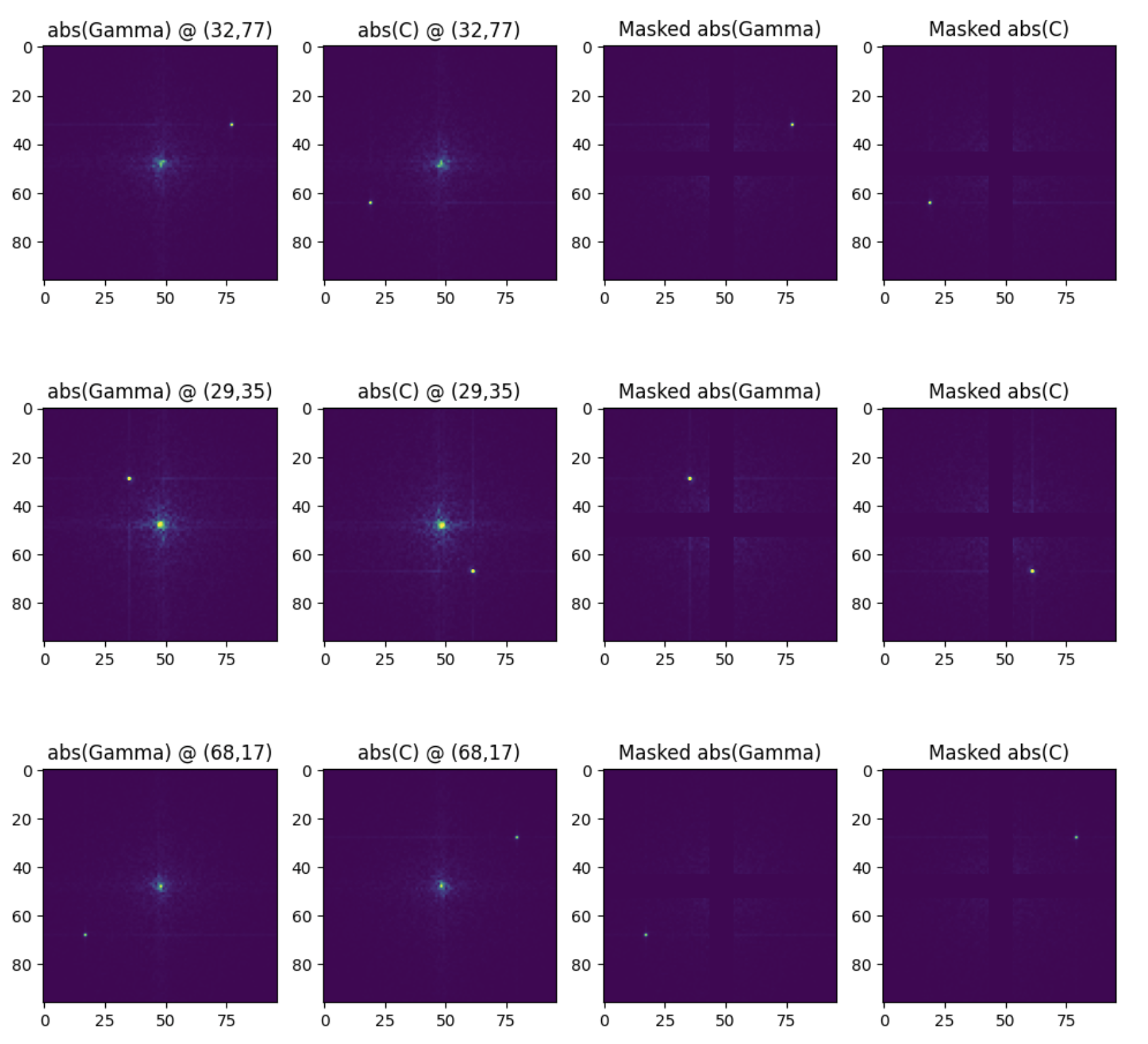}
    \caption{The Covariance Tensor of the image set in Fourier Space, plotted for four randomly chosen points. The first column shows $\Gamma$ and the second $C$. The second two columns show $\Gamma$ and $C$ with the central cross region masked out. The plots reveal the expected relation between the two covariances and a surprising diagonal structure for $\Gamma$.}
    \label{ComplexCovariance}
\end{figure}

We visualize the complex covariance in Fourier space similar to how we visualized the real covariance. In Figure \ref{ComplexCovariance} we have selected a few points at random and plotted $\Gamma$ and $C$. The plots are somewhat convoluted by the pixels in the region of the central cross so we masked these in the second two columns. These plots reveal a surprising structure. 

First, the plots of $\Gamma$ compared to $C$ simply verifies the relation (\ref{GammaCRelation}) --- the two plots are images of each other under a rotation by $180^\circ$. Moreover, the values of $\Gamma$ and $C$ are real, and they are approximately zero everywhere except at the position of the chosen coordinates for $\{kl\}$. This can be summarized succinctly by
\beqa
{C^{ij}}_{kl} &\approx & {\Gamma^{ij}}_{\wt{k}\wt{l}} \\
{\Gamma^{ij}}_{kl} &\approx & \Gamma^{ij}_{diag}\,\delta^{ij}_{kl}
\eeqa
where $\Gamma^{ij}_{diag}$ is a real function representing components along the diagonal, and $\delta^{ij}_{kl}=\delta^i_k\delta^j_l$. Thus, the Fourier transform has diagonalized the covariance tensor (at least in an approximate sense). 

This is a remarkable property that is deserving of emphasis. Of all of the transformations that we might have imagined to diagonalize the covariance tensor, it is the humble Fourier transform that actually does the job. The implication is that while neighboring points in real space are statistically correlated, by contrast in Fourier space each wave-vector $\vec{k}$ is statistically independent of every other wave vector. So in a sense we have isolated the discrete bits of information that encode the low-order statistical properties of the image set.

\begin{figure}[htbp]
    \centering
    \includegraphics[width=0.8\linewidth]{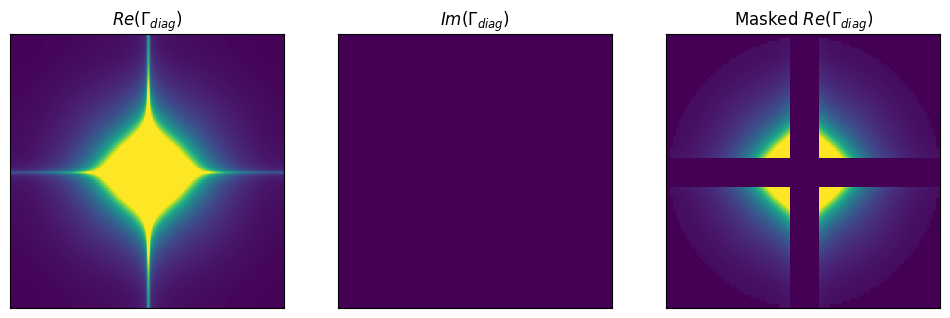}
    \caption{Here we plot the diagonal component $\Gamma^{ij}_{diag}$ of the covariance ${\Gamma^{ij}}_{jk}$ in Fourier Space. With the central cross masked, the remaining functional dependence is close to radial and falls off with distance.}
    \label{Gamma_diag}
\end{figure}

\begin{figure}[htbp]
    \centering
    \includegraphics[width=1.0\linewidth]{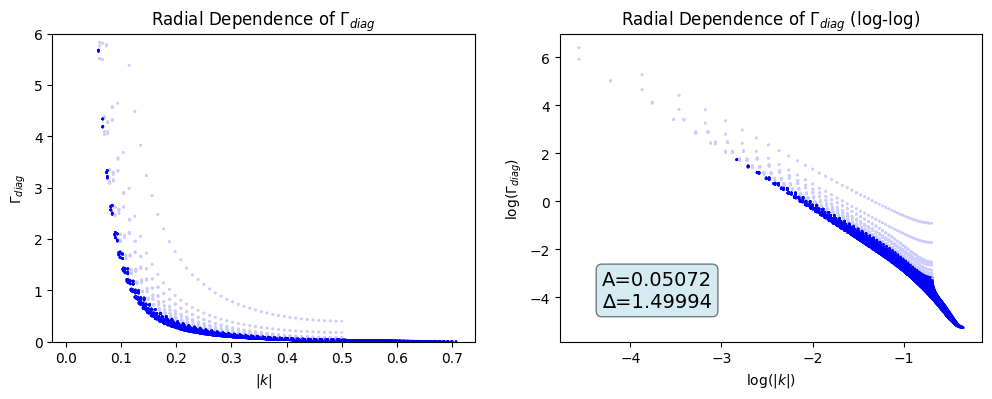}
    \caption{The dependence of $\Gamma_{diag}$ as a function of $|k|$ is plotted on a linear scale (left) and on a log-log scale (right). Points in the masked region of Figure \ref{Gamma_diag} are shown in light blue. A linear fit in log-log scale yields the parameters $A$ and $\Delta$ of the power-law covariance (\ref{PowerLawCovariance}).}
    \label{LinearFit}
\end{figure}

We now turn to the diagonal function, $\Gamma^{ij}_{diag}$. It contains the non-trivial content of the covariance in Fourier space. Recall that the diagonal of the covariance is the square of the point-wise standard deviation. We plot the real and imaginary parts as an image in Figure \ref{Gamma_diag}. 
The plots reveal that the covariance is real valued, and it has a distinctive, almost radial structure. If we ignore the central cross, which we do by applying a mask in the last column, the plot is highly symmetric. Specifically, the diagonal covariance is close to radial:
\beq
{\Gamma^{ij}}_{kl} \approx \Gamma^{ij}_{diag}\,\delta^{ij}_{kl} \quad\quad \Gamma^{ij}_{diag}\approx \Gamma^{ij}_{diag}(|k|)\,.
\eeq
To analyze the radial behavior, for each pixel we plot $\Gamma^{ij}_{diag}$ as a function of the radial distance $|k|$. The radial behavior is easiest to understand when plotted on a log-log scale as in Figure \ref{LinearFit}. 
In this figure the light blue points are the masked points in the central cross region. The remaining points exhibit a strikingly linear trend, and we do a linear fit on these points. The linear fit reveals the true structure of the covariance tensor. Inverting the logs and inserting the linear fit parameters gives:
\beq
{\Gamma^{ij}}_{kl} \approx \frac{A^2}{\lvert k\rvert^{2\Delta}} \delta^{ij}_{kl} \quad\quad \Delta \approx 1.5 \label{PowerLawCovariance}
\eeq
where $A$ is a constant that doesn't matter much (in fact we will alter it later). The important parameter is $\Delta$ and we will refer to it as the \textbf{Scaling Parameter}. 

This inverse power-law behavior of the covariance tensor places the image set within an important class of distributions that we will explore next.

\subsection{Modeling the Image Set as a Scale Invariant Distribution}
We saw above that the covariance tensor in Fourier space is diagonal and has the power law behavior given by (\ref{PowerLawCovariance}). This places the image set within a class of ``scale-invariant" distributions, so-called because they have certain properties that are invariant or self-similar under changes in scale. To see this, we define the dilation operator that acts on position vectors $\vec{x}$ by a simple rescaling: 
\beq
\phi_\lambda(\vec{x}) = \lambda \,\vec{x}\,.
\eeq
The wave-vector in Fourier space is inversely proportional to its corresponding wavelength, so it behaves inversely:
\beq
\phi_\lambda(\vec{k}) = \frac{\vec{k}}{\lambda}\,.
\eeq

Because of its power-law form (\ref{PowerLawCovariance}) the covariance in Fourier space is in fact an eigenstate of dilations with eigenvalue $\lambda^{2\Delta}$:
\beq
\phi_\lambda({\Gamma^{ij}}_{kl}) = \lambda^{2\Delta}\,{\Gamma^{ij}}_{kl}\,.
\eeq
Thus, under dilations, the covariance tensor changes amplitude but remains functionally the same. Probability Distributions with this property are referred to as \textbf{Scale-Invariant Distributions}.

An idealized distribution exhibiting this type of scale invariance is a multivariate normal distribution with covariance as described. For some values of the scale factor $\Delta$, these distributions have special names as summarized in Table \ref{ScaleInvariantNoise}.  
\begin{table}[htbp]
\centering
\begin{tabular}{|c|c|}
\hline
\multicolumn{2}{|c|}{\bf{Scale Invariant Noise}} \\
\hline\hline
Scale Factor & Name \\
\hline
$\Delta=0$ & White Noise \\
$\Delta=1$ & Pink Noise \\
$\bm{\Delta\approx 1.5}$ & \textbf{``Cloud Noise"} \\
$\Delta=2$ & Red (Brownian) Noise \\
\hline
\end{tabular}
\caption{Types of scale-invariant noise distributions.}
\label{ScaleInvariantNoise}
\end{table}
Noise with $\Delta=0$ is referred to as \textbf{White Noise} and it is the only truly scale {\it invariant} noise distribution as its covariance is properly invariant under dilations. For scale factor $\Delta=2$, the noise distribution is referred to as \textbf{Red Noise}. The term ``red" does not refer to color, but to the exaggeration of the amplitude of the long-wavelength, low-frequency modes (which would be on the red side of the rainbow if the wavelengths represented the visible color spectrum). To add to the confusion, Red Noise is often referred to as \textbf{Brownian Noise}, and sometimes shortened to Brown Noise. The term {\it Brown} here does not extend the color analogy; rather it refers to a person, Robert Brown, who observed the chaotic motion of small particles suspended in fluids, a phenomenon now known as Brownian Motion. Noise with scale factor $\Delta=1$ is referred to as \textbf{Pink Noise} because it is between White Noise and Red Noise. And between Pink Noise and Red Noise with $\Delta\approx 1.5$ we have \textbf{Cloud Noise}. This is not standard name, but a name we are coining here because noise of this type looks like...clouds! To see this, we need to find a way to generate the noise.

\begin{figure}[htbp]
    \centering
    \includegraphics[width=0.8\linewidth]{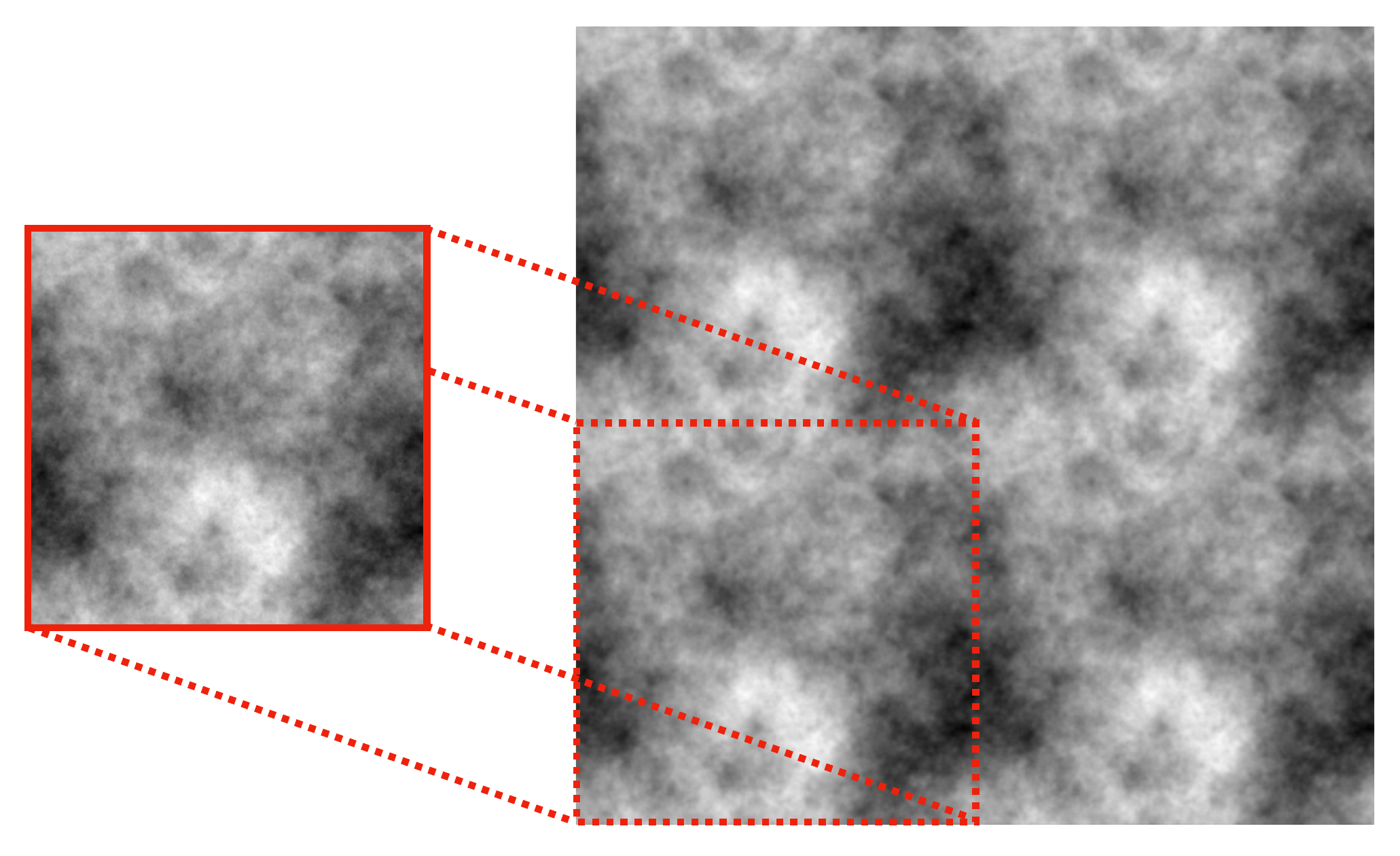}
    \caption{The naive procedure for generating scale invariant noise produces noise with periodic boundary conditions. This is shown here by tiling the noise in a wallpaper pattern. Note that the image is continuous and smooth on the edges where the tiles meet.}
    \label{PeriodicNoise}
\end{figure}

To create scale invariant noise images, we exploit some properties of white noise. White noise with zero mean and unit standard deviation is a sample from the normal distribution $\mathcal{N}(0,I)$. Its covariance tensor in Fourier space is ${\Gamma^{ij}}_{kl} \sim \delta^{ij}_{kl}$. Notice that this is diagonal and each component is independent of the frequency. Thus, white noise in real space can be thought of as an  independent normal distribution for each pixel. Unique to white noise, in Fourier space it also has the form of an independent normal distribution at each pixel. In fact, we can use the unique properties of white noise to construct any scale invariant noise profile. Here's the basic outline of the procedure: take white noise, Fourier transform it, divide each wave-mode by $|k|^{\Delta}$, and then inverse Fourier transform back to real space. Schematically this looks like the following:
\beq
\text{Scale Invariant Noise} = \mathcal{F}^{-1}\left[\frac{\mathcal{F}\left[\text{White Noise}\right]}{|k|^{\Delta}}\right]\,.
\eeq
This procedure comes close to generating the noise that we want, but there are subtleties. Upon close inspection, we notice that the procedure generates noise images with periodic boundary conditions as shown in Figure \ref{PeriodicNoise}. 
Specifically, the periodicity is toroidal, meaning that the top edge matches smoothly with the bottom edge, and the left edge matches smoothly with the right edge. When toroidally periodic noise is used in a diffusion model, it can lead to unwanted edge artifacts as the model eventually learns that the boundaries match smoothly.

\begin{figure}[htbp]
    \centering
    \includegraphics[width=0.8\linewidth]{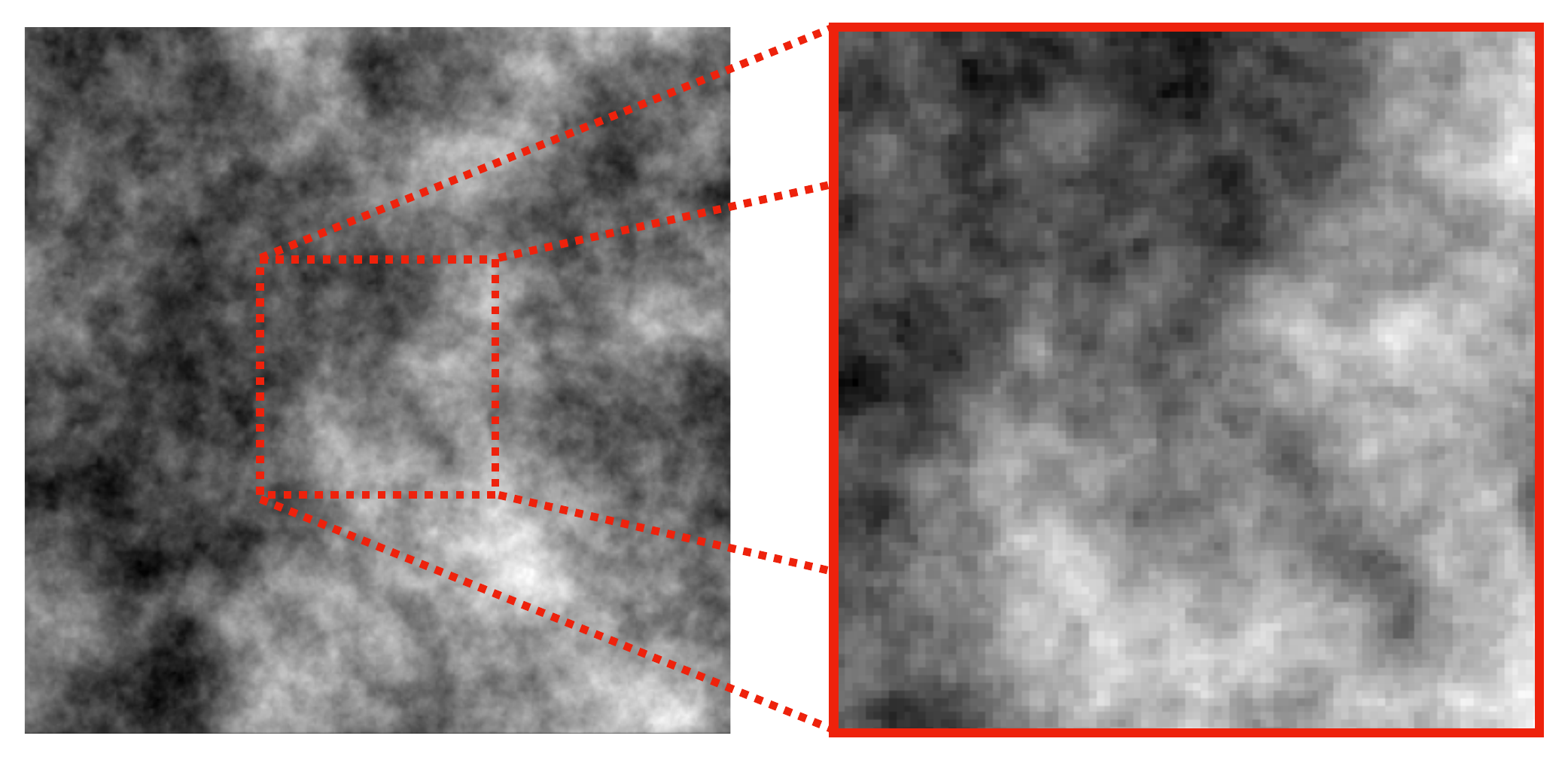}
    \caption{Producing periodic noise on a larger scale and cropping the center produces scale-invariant noise with aperiodic boundary conditions.}
    \label{CroppedNoise}
\end{figure}

We will use a trick to get around this. The trick is to create a noise image using the above procedure that is larger than the image we need, then we crop out the center of the image as shown in Figure \ref{CroppedNoise}. 
The resulting noise image will have the same scaling parameter. How do we know this? Because of scale invariance! Scale invariant noise is self-similar at every scale, so you can crop out any section of the image to get another noise image with the same scaling properties. You will need to readjust the overall amplitude by multiplying by a constant in order to match the standard deviation you want, but the scaling parameter $\Delta$ will be unchanged. In practice we've found that tripling the linear dimensions of the noise image and cropping out the center works reasonably well to remove the boundary periodicity. The amplitude $A$ in the covariance ${\Gamma^{ij}}_{kl}=\frac{A^2}{|k|^{2\Delta}}\delta^{ij}_{kl}$ is then adjusted after the cropping procedure to get a noise distribution with unit standard deviation.

\begin{figure}[htbp]
    \centering
    \includegraphics[width=0.9\linewidth]{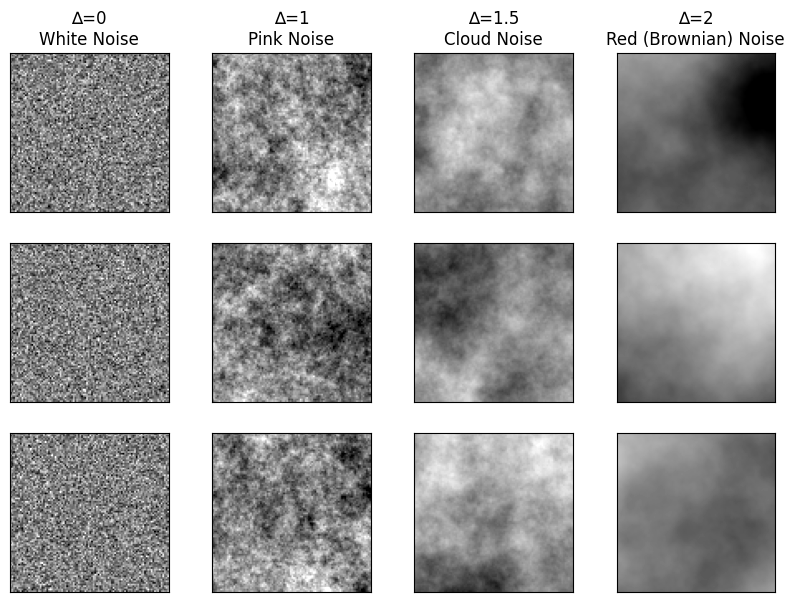}
    \caption{Noise samples for White Noise, Pink Noise, Cloud Noise, and Red (Brownian) Noise. Increasing $\Delta$ amplifies low-frequency modes and suppresses high-frequency modes. Cloud Noise has a scaling parameter tuned to the image dataset.}
    \label{NoiseSamples}
\end{figure}

Samples from the noise distributions for White Noise, Pink Noise, Cloud Noise, and Red (Brownian) Noise are shown in Figure \ref{NoiseSamples}. 

\subsection{What to Wager?}
Let's now come back to the hypothetical bet with the shady gambler. How do we use this information to guess the value of the second pixel? And how much do we wager for each guess? You now have some tools at your disposal to answer these questions. You've captured the low-order statistical properties of the image set using the mean and covariance. And you now have a noise generator for scale invariant noise that models these low-order statistical properties in an easily reproducible way. You can use this to guess the value of the second pixel for each image and propose a wager.

Is this the best you can do? Could you improve the statistical model? Likely, yes. You might try going to higher order to get better predictions for the statistical correlations. For example, the fourth order correlation known as {\it kurtosis} measures the non-Gaussianity of the distribution. This has been studied for natural image sets and it was found that they exhibit predictable kurtosis \cite{srivastava2003advances, ZoranKurtosis}.

But, we won't continue to improve the model here and craft a better strategy for the bet. Instead, we will use the statistical model we have constructed to improve diffusion models for image generation!

\section{Cloud Diffusion: Overview}
\begin{figure}[htbp]
    \centering
    \includegraphics[width=0.8\linewidth]{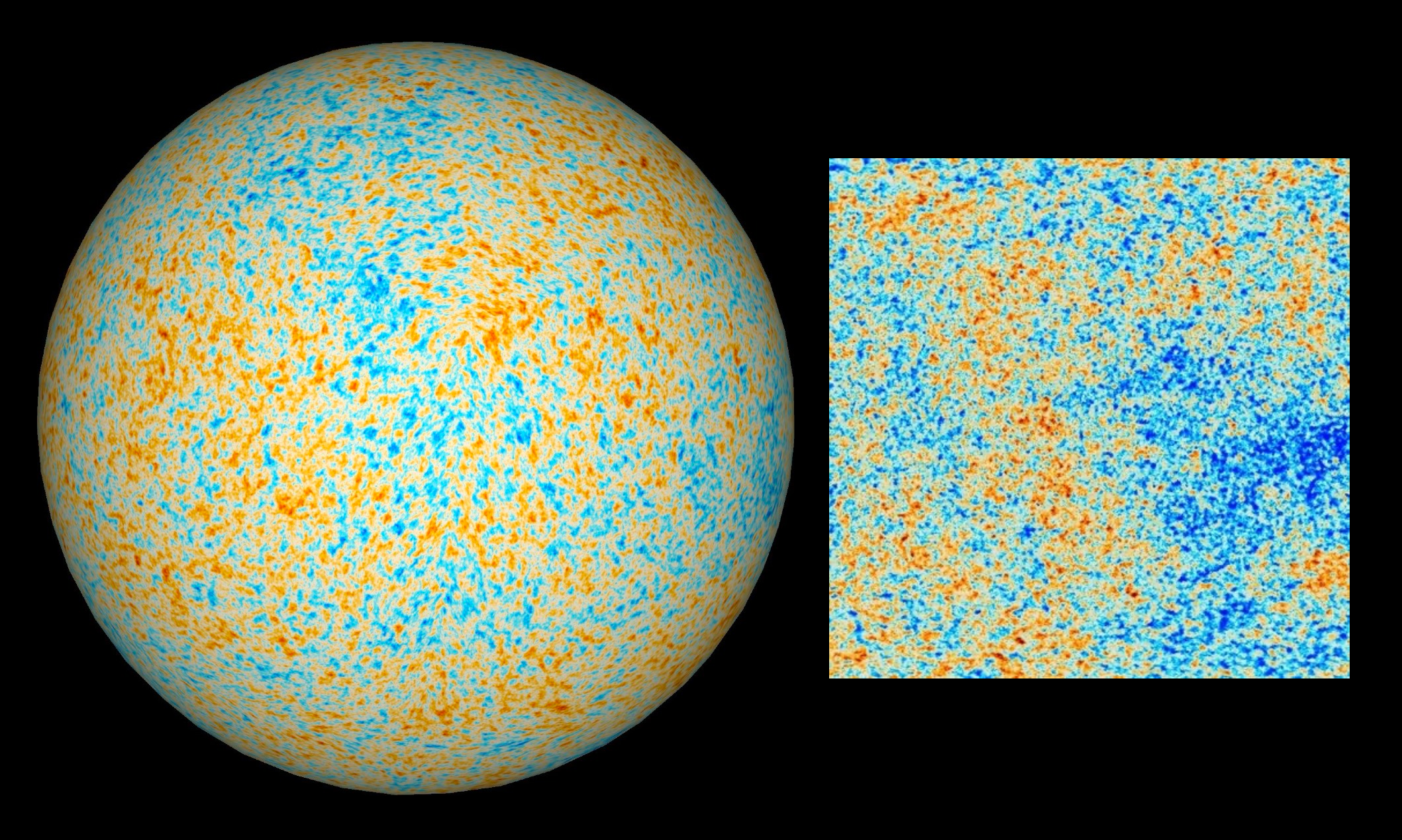}
    \caption{The density fluctuations of the Cosmic Microwave Background radiation, measured by the Planck mission, exhibit a scale-invariant power spectrum similar to that of Cloud Noise \cite{IPAC, CMB}.}
    \label{CMB}
\end{figure}

The universe was born from scale-invariant noise. Shortly after the big bang, the universe was an extremely uniform soup of fundamental particles. Small fluctuations in the energy density exhibited a scale-invariant power-spectrum as evidenced by the temperature fluctuations of the Cosmic Microwave Background radiation as shown in Figure \ref{CMB}. As the universe expanded and cooled, through gravitational collapse these small fluctuations condensed into the stars, galaxies, and clusters we see today. Diffusion models have a considerably less ambitious task: to create desirable images from pure noise. Nevertheless, we claim here that diffusion models can function better if they are seeded by the same type of scale-invariant probability density fluctuations.

\begin{figure}[htbp]
    \centering
    \includegraphics[width=1.0\linewidth]{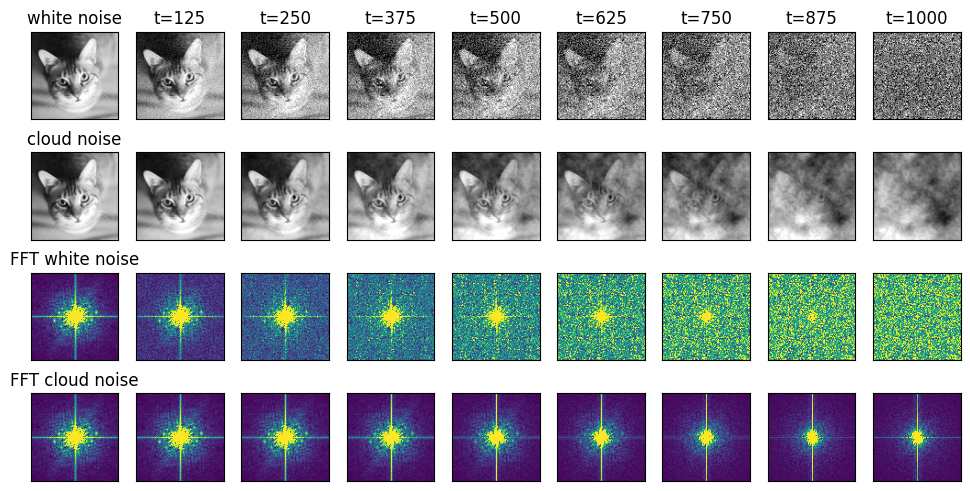}
    \caption{Here we visualize the diffusion noising procedure for white noise and for cloud noise. The noise schedule is shown in real space and in Fourier space where the absolute value is taken.}
    \label{WhiteAndCloudSamples}
\end{figure}

The fundamental premise behind Cloud Diffusion is that the diffusion paradigm can be improved if we replace white noise in the noising and denoising procedure with Cloud Noise tuned to the scaling parameters of the image dataset. The noising procedure for white noise and Cloud Noise is visualized in Figure \ref{WhiteAndCloudSamples}. 
Alternatives to white noise in the forward diffusion process have been considered before (see for example \cite{bansal2023cold}). Cloud Diffusion models differ in that they employ noise that is fine-tuned to match certain statistical properties of the image sets they models. Since Cloud Noise shares the low-order statistical properties, the models begin with a multivariate normal distribution that is {\it closer} (in a quantifiable sense) to the image set than their white noise counterparts. This allows the model to build images more holistically by treating all frequencies on equal footing.

A quick glance at Figure \ref{WhiteAndCloudSamples} gives a teaser of what's in store. Comparing the noising procedure in Fourier Space for white noise (third row) versus the noising procedure for Cloud Noise (last row) reveals that while the former procedure produces a noisy image that looks nothing like the original image from the data set, the latter procedure produces a noisy image that often looks very similar (in Fourier Space) to the original image to the untrained eye. For this reason, we shouldn't expect that training a Cloud Diffusion Model will be easier than training a white noise model --- and it very likely could be considerably more difficult. The true benefits of the model emerge during inference.

\subsection{Noising Procedure: White Noise Diffusion}
The noising procedure for Cloud Diffusion models is similar to their white noise counterparts. A key property of scale invariant noise makes this possible so we will take the time to highlight it here. 

First, we recall how the procedure unfolds for white noise diffusion models. The image is gradually corrupted, usually over around the order of 1000 time steps, by adding noise in a linear fashion. We first normalize the images. For image sets with low image diversity (for example the CelebA or FashionMNIST image sets) it is common to normalize by the pixel-averaged mean and standard deviation, $\mu$ and $\sigma$. But, as we've seen, for highly diverse image sets it is common that the pixel-wise mean, $\mu^{ij}$, and standard deviation, $\sigma^{ij}$, are highly uniform across the image. In these cases we can normalize the images on a pixel-by-pixel basis:
\beq
x^{ij} \longrightarrow \frac{x^{ij}-\mu^{ij}}{\sigma^{ij}}\,.
\eeq
The normalized image set has zero mean and unit standard deviation at each pixel. 

Two-dimensional white noise with square dimension $N\times N$ is sampled from the normal distribution $\mathcal{N}(0,I)$, where $I$ is the $N\times N$ identity matrix. It has the unique feature that the sampled value of each pixel is statistically independent of every other point. The noising procedure for a single image adds a little bit of white noise, corrupting the image at each time step, until the image is entirely replaced by noise. For each time step $t-1\rightarrow t$, we sample noise from the normal distribution $\vep_{t-1:t}\sim \mathcal{N}(0,I)$ and add it to the image linearly as follows:
\beqa
x^{ij}_1 &=& \sqrt{\alpha_{1}}x^{ij}_{0} + \sqrt{1-\alpha_1} \vep^{ij}_{0:1} \nn\\
x^{ij}_2 &=& \sqrt{\alpha_{2}}x^{ij}_{1} + \sqrt{1-\alpha_2} \vep^{ij}_{1:2} \nn\\
& \vdots & \nn\\
x^{ij}_t &=& \sqrt{\alpha_{t}}x^{ij}_{t-1} + \sqrt{1-\alpha_t} \vep^{ij}_{t-1:t}\,.\label{NoisingProcedure}
\eeqa
One way to view the corruption procedure is to think of the single image $x^{ij}_t$ as a sample from the normal distribution $\mathcal{N}(\sqrt{\alpha_{t}}x^{ij}_t,(1-\alpha_t)I)$. While this is accurate, for our purposes it is more useful to consider the corrupted image set at each time step as a probability distribution $\mathcal{Q}_t$ and consider its aggregate statistics. From this perspective, the coefficients $\sqrt{\alpha_t}$ and $\sqrt{1-\alpha_t}$ are chosen so that $\mathcal{Q}_t$ at each time step retains the same unit standard deviation at each pixel. Said another way, while $\mathcal{Q}_t$ at each time step $t$ is not a normal distribution, and its variance ${\Sigma^{ij}}_{kl}(t)$ is not constant over time, the diagonal component of the covariance, $\Sigma^{ij}_{diag}={\Sigma^{ij}}_{ij}$, is constant, and it is equal to $1$ at each point.

The procedure can be drastically simplified by what may be called a {\it jump trick}. First note that the cumulative noise added at time $t$ is a sum of the small amount of noise added at each time step. The trick relies on the property that can be represented schematically as
\beq
\text{White Noise } + \text{ White Noise } = \text{ White Noise}\,.
\eeq
This derives from an addition property of normal distributions. In fact, there is a more general addition formula for any two multivariate normal distributions with {\it diagonal} covariance tensors. Consider two such distributions $\mathcal{Q}_1 = \mathcal{N}(\bm{\mu_1}, \bm{\sigma^2_1}I)$ and $\mathcal{Q}_2 = \mathcal{N}(\bm{\mu_2}, \bm{\sigma^2_2}I)$. If $x\sim \mathcal{Q}_1$ and $y\sim \mathcal{Q}_2$, then the sum $x+y$ is a sample from the distribution $\mathcal{Q}_3 = \mathcal{Q}_1\oplus \mathcal{Q}_2$. The summation formula
\beq
\mathcal{N}(\bm{\mu_1}, \bm{\sigma^2_1}I) \oplus \mathcal{N}(\bm{\mu_2}, \bm{\sigma^2_2}I) = \mathcal{N}((\bm{\mu_1}+\bm{\mu_2}), (\bm{\sigma^2_1}+\bm{\sigma^2_2})I) \label{NormalSummation}
\eeq
reveals that the resultant distribution $\mathcal{Q}_3$ is itself a multivariate normal distribution. And furthermore, it is also diagonal. Given a sample from an arbitrary multivariate normal distribution, $x\sim \mc{N}(\bm{\mu}, \bm{\Sigma})$, the rescaling $x\rightarrow \alpha x$ by a real parameter $\alpha$ is equivalent to sampling from a rescaled normal distribution denoted $\alpha\,\mc{N}(\bm{\mu}, \bm{\Sigma})$. The rescaled normal distribution satisfies
\beq
\alpha\,\mc{N}(\bm{\mu}, \bm{\Sigma}) = \mc{N}(\alpha \bm{\mu}, \alpha^2\bm{\Sigma})\,. \label{RescaledNormal}
\eeq
Putting (\ref{NormalSummation}) and (\ref{RescaledNormal}) together gives a general formula for the linear combination of two diagonal normal distributions:
\beq
\alpha\,\mathcal{N}(\bm{\mu_1}, \bm{\sigma^2_1}I) \oplus \beta\,\mathcal{N}(\bm{\mu_2}, \bm{\sigma^2_2}I) = \mathcal{N}\left((\alpha\bm{\mu_1}+\beta\bm{\mu_2}), (\alpha^2\bm{\sigma^2_1}+\beta^2\bm{\sigma^2_2})I\right)\,. \label{LinearNormal}
\eeq

Using these properties, it is straightforward to show that 
\beqa
x^{ij}_t = \sqrt{\bar{\alpha}_t} x^{ij}_0 + \sqrt{1-\bar{\alpha}_t}\vep^{ij}_{0:t}
\eeqa
where $\vep^{ij}_{0:t}\sim \mathcal{N}(0,I)$ is a white noise sample and 
\beq
\bar{\alpha}_t = \alpha_1\alpha_2\dots \alpha_t \,.
\eeq
We call this a {\it jump trick} because it allows us to jump straight from the original image at $t=0$ to a corrupted image at any time $t$ without going through the process of adding a small amount of noise at each step. The noise $\vep^{ij}_{0:t}$ can be thought of as the cumulative noise acquired over all time steps from $0$ to $t$. But the nice properties of normal distributions under linear combination imply that it is itself a sample from a normal distribution. 

This turns out to be a critical property that the diffusion architecture relies on. Not only does it allow us to easily generate noisy images at any timestep, the diffusion model uses it implicitly in the training procedure --- the model is actually trained to guess the cumulative noise tensor $\vep^{ij}_{0:t}$ for a corrupted image at any given time step.

It is sometimes convenient to re-parameterize this expression using angular variables. Setting $\cos\theta_t=\sqrt{\alpha_t}$ and $\sin\theta_t=\sqrt{1-\alpha_t}$ we have
\beq
x^{ij}_t = \cos\theta_t \,x^{ij}_0 + \sin\theta_t\,\vep^{ij}_{0:t}\,.
\eeq
This gives us a nice conceptual picture of the noising procedure as slowly rotating the images into a distribution of white noise. 

\subsection{Noising Procedure: Cloud Diffusion}
Cloud diffusion works on the same principles, but with cloud noise in place of white noise. For the model to work as intended, we need to show that there is an analogous summation formula, given schematically as 
\beq
\text{Cloud Noise } + \text{ Cloud Noise } \overset{\textbf{?}}{=} \text{ Cloud Noise}\,.
\eeq
And specifically, we need to show that the scaling parameter $\Delta$ is preserved in the addition formula (though the magnitude of the covariance may change). The fact that Cloud Noise in diagonal in Fourier space gives us hope that the normal distribution summation formula (\ref{LinearNormal}) may apply. However, the situation is complicated because we are dealing with {\it complex} normal distributions in Fourier space.

\begin{figure}[htbp]
    \centering
    \includegraphics[width=0.6\linewidth]{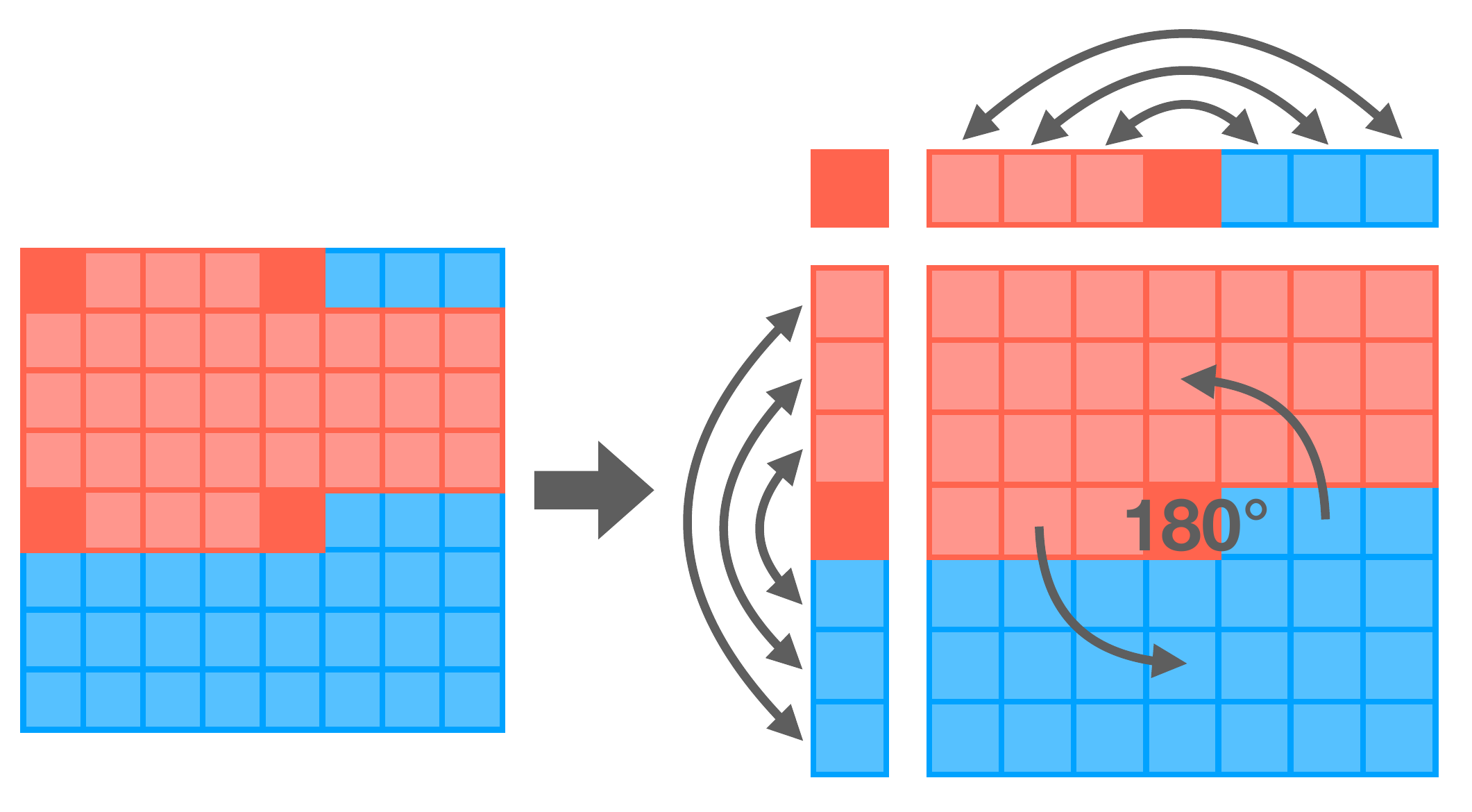}
    \caption{For square tensors with even linear dimensions, the origin of the Fourier transform is necessarily off center, and the involution is not a simple rotation. Instead the left column and top row invert as depicted, while the bottom right square simply rotates $180^\circ$.}
    \label{Inversion}
\end{figure}

Rather than perform the analysis on the complex normal distributions, we will take a different approach here. As we have seen, since we are dealing with real images, when projected into Fourier space, the images $Z^{ij}=\mc{F}(x^{ij})$ are constrained by a relation under an involution represented by a mapping of coordinate $(i,j)\rightarrow (\wt{i},\wt{j})$. For square $N\times N$ arrays with odd $N$, this involution is simply a $180^\circ$ rotation about the center pixel. However, when $N$ is even there are subtleties in the implementation of the Fast Fourier Transform. First note that we have shifted the center point of the FFT to the center of the tensor (in PyTorch this is accomplished by torch.fft.fftshift). However, for even dimensions the center point is shifted one pixel right and one pixel down, so the inversion is not a simple rotation. Instead the left column and top row are treated separately, inverting pixels within themselves, while the remaining bottom right square array simply inverts by a $180^\circ$ rotation. The full inversion is pictured in Figure \ref{Inversion}.
With the inversion properly defined, for all square tensors the Fourier transform of a real image satisfies the involution relation $Z^{\wt{i}\wt{j}}=(Z^{ij})^*$. This effectively eliminates half of the complex degrees of freedom so that they match the degrees of freedom of the real images. In turn the relation constrains the covariance tensors by ${C^{ij}}_{kl}={\G^{ij}}_{\wt{k}\wt{l}}$. This begs the question, can we do all of this in the real domain and avoid complex numbers altogether? In fact, we can. While PyTorch has methods that purport to isolate the true degrees of freedom (e.g. torch.fft.rfft2), in practice one finds that these methods usually still return a complex tensor whose components are not all independent.

\begin{figure}[htbp]
    \centering
    \includegraphics[width=0.7\linewidth]{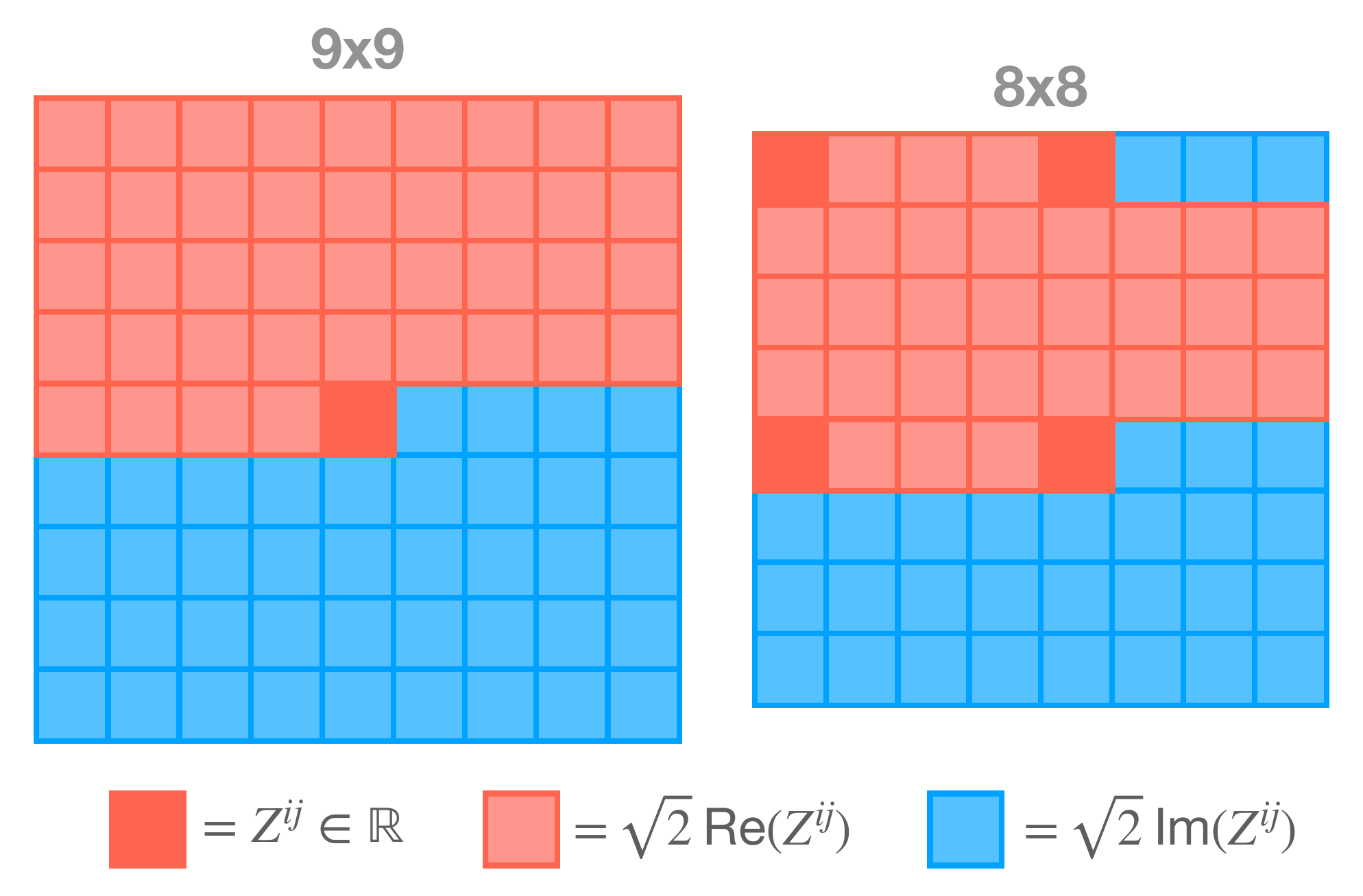}
    \caption{Under the involution, light red pixels map to light blue pixels. Solid red pixels are stationary, and they are real valued under the Fourier transform. The Real Fourier Transform takes the ordinary Fourier transform, then projects the light red squares into their real part, projects the light blue pixels into their imaginary part, and retains the value of the solid red pixels. }
    \label{RFFT}
\end{figure}

Our goal here is to find a modified Fast Fourier Transform $X^{ij}=\mathcal{F}_{\mathbb{R}}(x^{ij})$ of a real image that
\begin{enumerate}
\item{Is itself real-valued}
\item{Has the same information content as the ordinary 2D FFT}
\item{Returns a tensor of the same shape ($N\times N$ for square input)}
\item{Respects the power-law statistical properties of the image set.}
\end{enumerate}
Our strategy can be explained qualitatively as follows. Divide up $Z^{ij}=\mathcal{F}(x^{ij})$ into sectors that are images of each other under the involution taking $(i,j)\rightarrow (\tilde{i},\tilde{j})$. Combine the real part of one sector with the imaginary part of the other to form a real tensor as depicted in Figure \ref{RFFT}. The result is a real-valued tensor $X^{ij}=\Pi(Z^{ij})$ that has the same information content as $Z^{ij}$. Said another way, given $X^{ij}$ we can easily reconstruct $Z^{ij}$, meaning the projection is invertible so that $Z^{ij}=\Pi^{-1}(X^{ij})$. The Real Fourier Transform is then simply
\beqa
X^{ij} &=& \mc{F}_{\bb{R}}(x^{ij}) \\
&=& \Pi\left(\mc{F}(x^{ij})\right)
\eeqa
and its inverse is $\mc{F}^{-1}_{\bb{R}}(X^{ij}) = \mc{F}^{-1}\left(\Pi^{-1}(X^{ij})\right)=x^{ij}$. We also note that since the Real Fourier transform uses the real and imaginary parts, it is not linear over the complex domain. However, the transformed images are real and have the same dimension, so we have a map from a vector space isomorphic to $\bb{R}^{N^2}$ to another vector space, also isomorphic to $\bb{R}^{N^2}$. In the real domain, the map is linear.  

Since we are dealing with image sets where both the real and imaginary parts of $Z^{ij}$ each satisfy the same power-law statistics in aggregate, $X^{ij}$ will also have power-law statistics in aggregate. Consequently, denoting the Real Fourier Transform of the image distribution $\mc{Q}$ by $\wt{\mc{Q}}$, its covariance $\wt{\Sigma}{{}^{ij}}_{kl}$ is real, and is given by
% Note: the strange way that we represented indices here is because LaTeX is not formatting the expression properly when the symbol \Sigma is replaced with \widetilde{\Sigma}, giving rise to double subscript errors messages. This gets around that problem. 
\beq
\wt{\Sigma}{{}^{ij}}_{kl}\approx \frac{A^2}{|k|^{2\Delta}}\,\delta^{ij}_{kl}\,.
\eeq
This gives us another way of constructing scale-invariant noise. We take (real-valued) white noise of some larger square dimension than we need (square dimensions $3N\times 3N$ are usually sufficiently large), then multiply each point by $\frac{A}{|k|^{\Delta}}$, take the inverse Real Fourier Transform, and then excise the central $N\times N$ square. This is equivalent to the procedure we gave above. However, note that now, in the intermediate step we have a $3N\times 3N$ real distribution with a covariance tensor that is precisely diagonal and given by $\wt{\Sigma}{{}^{ij}_{}}_{kl} = \frac{A^2}{|k|^{2\Delta}}\,\delta^{ij}_{kl}$. From the addition of diagonal normal distributions formula (\ref{LinearNormal}), the sum of two such distributions has the same power-law scaling:
\beq
\alpha\,\mc{N}\left(0, \frac{A^2}{|k|^{2\Delta}} I\right) \oplus \beta\,\mc{N}\left(0, \frac{A^2}{|k|^{2\Delta}} I\right) = \mc{N}\left(0, \frac{(\alpha^2 + \beta^2)A^2}{|k|^{2\Delta}} I\right) \label{SumOfDiagonalNormals}
\eeq
The overall amplitude of the covariance changed, but its power-law functional form is the same. Consequently, the sum is still scale-invariant noise with the same scaling parameter $\Delta$.

This is great news, and it means that we can carry out the noising procedure for the Cloud Noise model in the exact same way we did for the white noise model. Since the Real Fourier Transform is linear over the reals, this can be done in real space, or in Real Fourier Space. We still progressively add noise over many time steps, which looks the same in real space as before:
\beqa
x^{ij}_1 &=& \sqrt{\alpha_{1}}x^{ij}_{0} + \sqrt{1-\alpha_1} \vep^{ij}_{0:1} \nn\\
x^{ij}_2 &=& \sqrt{\alpha_{2}}x^{ij}_{1} + \sqrt{1-\alpha_2} \vep^{ij}_{1:2} \nn\\
& \vdots & \nn\\
x^{ij}_t &=& \sqrt{\alpha_{t}}x^{ij}_{t-1} + \sqrt{1-\alpha_t} \vep^{ij}_{t-1:t}\,.
\eeqa
The only difference is that now at each time step $\vep^{ij}_{t-1:t}$ is a sample from the Cloud Noise distribution. We have just shown that all scale-invariant noise, including Cloud Noise, satisfies the same rules (\ref{SumOfDiagonalNormals}) under linear combinations. Thus, we can perform the same {\it jump trick}:
\beqa
x^{ij}_t = \sqrt{\bar{\alpha}_t} x^{ij}_0 + \sqrt{1-\bar{\alpha}_t}\vep^{ij}_{0:t}
\eeqa
where $\bar{\alpha}_t = \alpha_1\alpha_2\dots \alpha_t$. And this can be sinusoidally-parameterized:
\beq
x^{ij}_t = \cos\theta_t \,x^{ij}_0 + \sin\theta_t\,\vep^{ij}_{0:t}\,.
\eeq

\subsection{White Noise Diffusion vs. Cloud Noise Diffusion\label{WhiteVsCloud}}
\begin{figure}[htbp]
    \centering
    \includegraphics[width=0.9\linewidth]{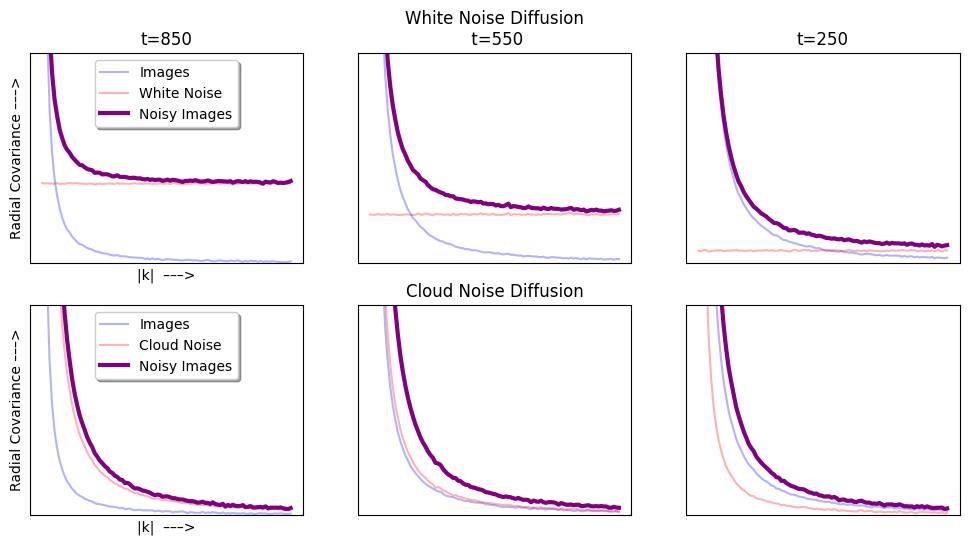}
    \caption{The reverse diffusion procedure for white noise and Cloud Noise. Removing white noise effectively exposes higher frequencies. By contrast, the signal to noise ratio increases uniformly across all frequencies as Cloud Noise is removed.}
    \label{CovarianceThroughTime}
\end{figure}

Everything looks very similar, but there are key differences. Consider the covariance tensors for the white noise diffusion model and the Cloud Diffusion model. For the white noise model, the noisy distribution $\mc{Q}_t$ has a covariance tensor ${{\Sigma_t}^{ij}}_{kl}$ that is not constant in time. The diagonal components $\Sigma^{ij}_{diag} = {{\Sigma_t}^{ij}}_{ij}$ are constant and equal to $1$ for all $ij$ since the images and noise are normalized and the coefficients $\sqrt{\alpha_t}$ and $\sqrt{1-\alpha_t}$ were chosen judiciously. But the off-diagonal components decrease as the images become noisier. In Real Fourier Space, we can calculate the approximate covariance tensor as the sum of that for the image set and for white noise:
\beq
{\wt{\Sigma}_t}{{}^{ij}}_{kl} \approx \cos^2\theta_t \frac{A^2}{|k|^{2\Delta}}\delta^{ij}_{kl} +\sin^2\theta_t \, \delta^{ij}_{kl}\,.
\eeq
By contrast, since the Cloud Noise distribution has been tuned to the scaling parameters of the dataset, the covariance tensor is the same at each time step. Namely, for all time steps, the covariance of $\wt{\mc{Q}}_t$ is
\beq
{\wt{\Sigma}_t}{{}^{ij}}_{kl} \approx \frac{A^2}{|k|^{2\Delta}}\delta^{ij}_{kl} \quad \text{for all timesteps } t\,.
\eeq
This has important implications as it concisely encapsulates the advantages of Cloud Diffusion Models over white noise diffusion models. The effect on the radial component of the covariance is shown for white noise and Cloud Noise diffusion in Figure \ref{CovarianceThroughTime}. 
Progressing in reverse timestep order, for white noise the effect of removing noise is to distill higher and higher frequency ranges. The white noise procedure obscures high frequencies (where the signal to noise ratio is low) while exposing low frequencies (where the signal to noise ratio is high). The exposed/obscured dividing line depends on time, and shifts toward higher frequencies as noise is removed. As seen in Figure \ref{CovarianceThroughTime} the dividing line is a loose boundary centered roughly on the intersection point of the image set covariance and the noise covariance. This manifests in the inference procedure as sequential image creation from low frequencies to high frequencies. Image generation occurs over many time steps, and the image starts out blurry with blobs of color and value, and gradually refines into a higher resolution image as the high frequency details emerge. This sequential inference procedure and its relation to auto-regressive models was discussed in a recent blog post \cite{DielemanBlogPost}. 

By contrast, for the Cloud Noise diffusion procedure the signal to noise ratio is the same for all frequencies. In this sense, it treats all frequencies on the same footing. Removing noise simply increases the overall signal to noise ratio. In the next sections we will show how this leads to advantages for Cloud Diffusion Models over white noise diffusion models. 

\section{Cloud Diffusion: Motivation}
Much of the advantage of the Cloud Diffusion architecture stems from the way it progressively builds images through the many time steps of the reverse diffusion process. As we have seen, white noise models build images in sequential order, distilling low frequency modes first and gradually progressing to higher frequency modes. By contrast, Cloud Diffusion Models build images more holistically, refining all frequencies simultaneously. This can have distinct advantages.

While there are many possible advantages to the Cloud Diffusion architecture, we focus on three broad categories here. We will argue that Cloud Diffusion Models can lead to
\begin{enumerate}
    \item {\bf Faster Image Generation}
    \item {\bf Better Details}
    \item {\bf Improved Conditional Guidance.}
\end{enumerate}
We expound the arguments for each of these bullet-points in the following sections.

\subsection{Faster Image Generation}
\begin{figure}[htbp]
    \centering
    \includegraphics[width=0.8\linewidth]{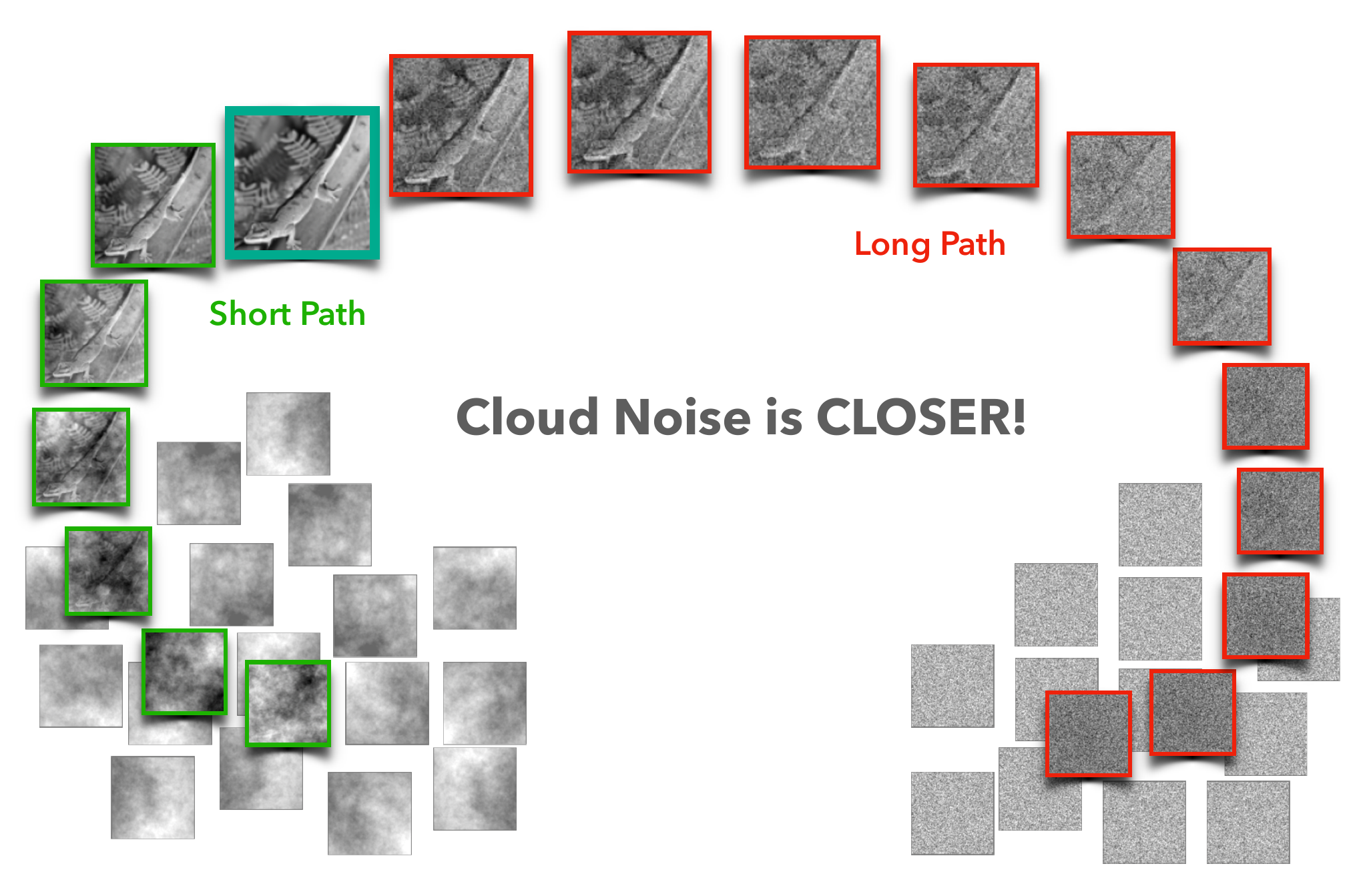}
    \caption{The reverse diffusion process carves a path from the noise distributions to the generated image set. We expect that since Cloud Noise has scaling parameters that are tuned to dataset, the path should be shorter in some quantifiable sense.}
    \label{DistributionPaths}
\end{figure}

The scaling parameter of scale-invariant Cloud Noise has been fine-tuned to match the low-order statistics of the image set. Thus, one might expect that Cloud Noise is in some sense ``closer" to the image set than white noise. The inference procedure can be thought of as carving out a path from the noise distribution to the image distribution. If Cloud Noise is closer to the image set than white noise, we should expect that the typical path from a point in the Cloud Noise distribution to a point in the image distribution will be shorter on average than the typical path from a point in the white noise distribution to a point in the image distribution as depicted in Figure \ref{DistributionPaths}. Shorter paths may equate to faster image generation. Can we quantify the distance between these distributions?

In fact, there is a natural measure for distance that can be defined on our distributions. To set up the problem, first consider a single variable normal distribution $\mc{Q}_0 = \mc{N}(\mu,\sigma^2)$. The probability density for the distribution is $\rho(x)\sim e^{-(x-\mu)^2/2\sigma^2}$. We can define the distance function $d(x,\mu|\mc{Q}_0) = \sqrt{(x-\mu)\sigma^{-2}(x-\mu)}$, which has the clear interpretation as a measure of the distance between the point $x$ and the center of the distribution $\mu$, scaled by the parameter $\sigma$. The probability density exponentially falls off with the square of the distance from the center
of the distribution $\rho(x)\sim e^{-d^2/2}$. We can extend this concept to a multivariate distribution $\mc{Q}_0=\mc{N}(\bm{\mu},\bm{\Sigma})$ by noting that the probability density is given by
\beq
\rho(\bm{x}) = \frac{\exp\left(-\frac{1}{2}{\color{red}(\bm{x}-\bm{\mu})^T\bm{\Sigma}^{-1}(\bm{x}-\bm{\mu})}\right)}{\sqrt{(2\pi)^n\det(\bm{\Sigma})}}\,.
\eeq
The expression in {\color{red} red} can be thought of as the square of a distance measure that has the same interpretation as the distance between the point $\bm{x}$ and the center of the distribution $\bm{\mu}$. The eigenvalues of $\bm{\Sigma}$ provide the scaling parameters of the distance function in each of the principle axes. In fact, this is a special case of a distance measure called the {\bf Mahalanobis Distance} that can be defined more generally \cite{Mahalanobis1936}. 

Let's relax the constraint that the distribution is normal, and simply consider a probability distribution $\mc{Q}_0$, which will serve as a reference distribution on which the measure is defined. Now consider two points, $x\sim \mc{Q}_1$ and $y\sim \mc{Q}_2$, sampled from two other distributions, which may be the same or different from $\mc{Q}_0$. The Mahalanobis distance between $x$ and $y$ relative to the reference distribution $\mc{Q}_0$ is given by
\beq
d_M(\bm{x},\bm{y}|\mc{Q}_0) = \sqrt{(\bm{x}-\bm{y})^T\bm{\Sigma}^{-1}(\bm{x}-\bm{y})}\,.
\eeq
It is common to take $\bm{y}$ to be the mean of $\mc{Q}_0$ as we did above, in which case we simply call it the Mahalanobis distance to the single point 
\beqa
d_M(\bm{x}|\mc{Q}_0) &\equiv & d_M(\bm{x},\bm{\mu}|\mc{Q}_0) \nn\\
&=& \sqrt{(\bm{x}-\bm{\mu})^T\bm{\Sigma}^{-1}(\bm{x}-\bm{\mu})}\,.
\eeqa
By taking expectation values, we can construct a measure of the distance between distributions:
\beq
d_M(\mc{Q}_1, \mc{Q}_2|\mc{Q}_0) \equiv \bb{E}_{\bm{x}\sim \mc{Q}_1}\bb{E}_{\bm{y}\sim \mc{Q}_2}\left[d_M(\bm{x},\bm{y}|\mc{Q}_0)\right]\,.
\eeq
Note that when the two distributions are the same, $\mc{Q}_1 = \mc{Q}_2$, the distance is not zero, but can be interpreted as a measure of the spread of the distribution. Similarly, we can define the distance between the single distribution $\mc{Q}_1$ and the center of the reference distribution $\mc{Q}_0$ by
\beq
d_M(\mc{Q}_1|\mc{Q}_0)\equiv \bb{E}_{\bm{x}\sim \mc{Q}_1}\left[d_M(\bm{x}|\mc{Q}_0)\right]\,.
\eeq

With these definitions, as promised we can quantify the notion of distance between white noise, Cloud Noise, and our dataset. We'll work in Real Fourier Space, and take the reference $\wt{\mc{Q}}_0$ to be the idealized distribution with covariance $\wt{\Sigma}{{}^{ij}}_{kl}=\frac{1}{|k|^{2\Delta}} \delta^{ij}_{kl}$. Since the covariance is diagonal, it is easy to invert: $\wt{\Sigma}^{-1} {{}^{ij}}_{kl}=|k|^{2\Delta} \delta^{ij}_{kl}$. The Mahalanobis distance is then
\beq
d_M(\bm{X},\bm{Y}|\mc{Q}_0) = \sqrt{(X_{ij}-Y_{ij})|k|^{2\Delta}(X^{ij}-Y^{ij})}
\eeq
noting that there is an implicit sum over repeated indices, $|k|$ is itself a function of $i$ and $j$, and $\bm{X}$ and $\bm{Y}$ are the transformed images in Real Fourier Space.

\begin{figure}[htbp]
    \centering
    \includegraphics[width=0.8\linewidth]{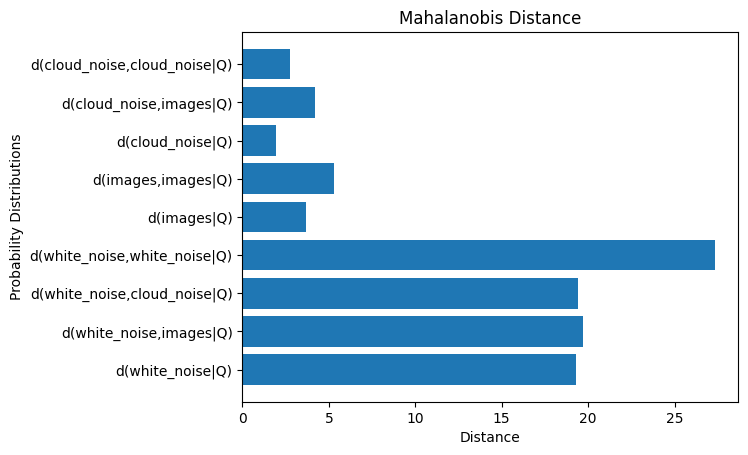}
    \caption{The Mahalanobis distances calculated between the various distributions as labeled. Note that Cloud Noise is clearly closer to the image set than white noise relative to this measure.}
    \label{MahaDistances}
\end{figure}

We can now take the expectation values to calculate the Mahalanobis distances among white noise, Cloud Noise, and the image set, all relative to our idealized reference distribution. The distances are plotted in Figure \ref{MahaDistances}.
The values from the figure quantify what we mean by {\it Cloud Noise is closer to the image set}. Shorter paths from the Cloud Noise distribution to the image set may translate into faster image generation if the inference procedure of the model is optimized appropriately. For example, we might expect that the Cloud Noise image generation procedure will produce high quality images with fewer time steps.

\subsection{Better Details}
\begin{figure}[htbp]
    \centering
    \includegraphics[width=1.0\linewidth]{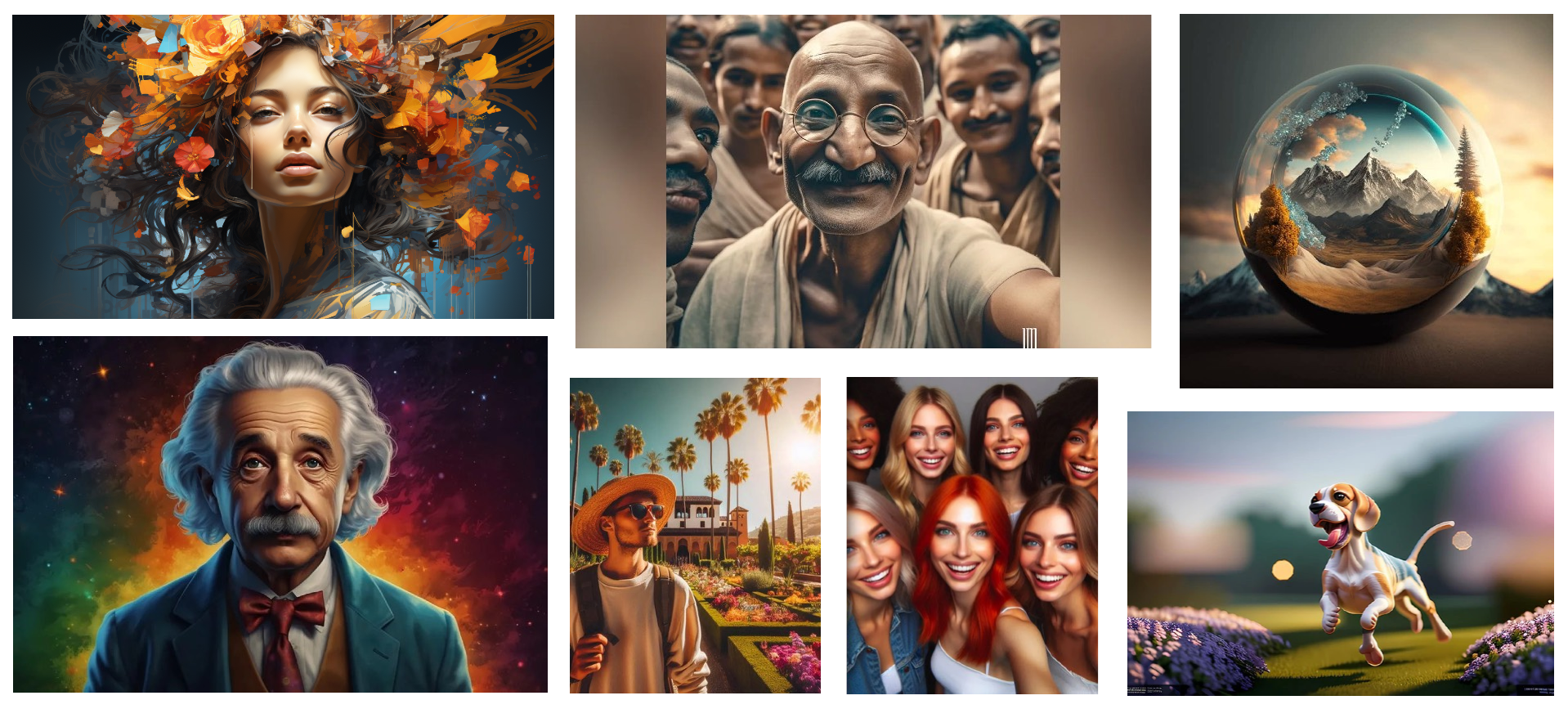}
    \caption{Images generated by diffusion models often exhibit a characteristic ``airbrushed" or ``ultra-processed" look, suggesting they poorly model the high frequency details of the image sets they are trained on.}
    \label{Airbrushed}
\end{figure}

We argue here that Cloud Diffusion can yield better high frequency fidelity to the image set that it is trained on. Images generated from classic, white noise diffusion models often have a characteristic look that might be described as ``squeaky-clean", ``ultra-processed", or (slightly anachronistically) ``airbrushed". The effect comes from the high-frequency details of the generated images, which often poorly model those of the image set the model was trained on. Textures are smoothed, and the gritty details that lend an image its realistic look are often filled in with gradients of color and value as seen in Figure \ref{Airbrushed}. 
There are multiple reasons for this. The most common explanations are the heavy use of upscaling to increase the resolution of the core model, and the reliance on a latent diffusion architecture that typically uses a Variational Auto-Encoder to compress the image to a latent space where the diffusion occurs followed by a decompression phase. 

We argue here that these are likely not the only reasons. Some of the high frequency fidelity loss can be attributed to the use of white noise in the noising procedure, and the sequential low-to-high frequency ordering of the inference procedure that it forces.

It has been observed since the early days of diffusion that the models build images in a sequential order, starting out as fuzzy blobs of color and value, and gradually refining as finer details are distilled. We explained the reasons for this in Section \ref{WhiteVsCloud}. For white noise models, the noising procedure effectively hides high frequencies while exposing low frequencies as shown in Figure \ref{CovarianceThroughTime}. The cutoff point between the frequencies that are exposed and the frequencies that are obscured shifts toward higher and higher frequencies as the noise is removed. The model can accurately predict the noise when the signal-to-noise ration is high, but fails when the ratio is too low. Thus, at any given timestep, the model can accurately predict the low-frequency components of the noise up to some cutoff beyond which the signal-to-noise ratio is too low for good predictions. Consequently, in the inference procedure, low frequencies are generated first, and higher frequencies are generated subsequently as the noise is progressively removed.

The ramifications of this sequential image building are twofold. First, the model only has a comparatively small window where it can distill high-frequency signal modes from the noise. For example, in a reverse-diffusion procedure with $1000$ time steps, the highest frequency modes may only be accessible after timestep $900$. Before that, the signal-to-noise ratio is too high for the model to accurately predict the noise (hence the predicted signal). This gives a wide range of time steps where the the low-frequency modes can be generated, but a comparatively narrow range of time steps where the high frequencies can be generated. Compounding this effect, the high-frequency range occupies a larger volume of the free-parameter space --- there are more high-frequency modes than low-frequency modes in Fourier Space. So narrowing the window where they can be generated is significantly constraining. 

Second, part of the magic behind the mechanism that makes diffusion models work is their Bayesian-esque approach to image building. The image builds over many time steps and gradually modifies itself. This inference procedure is guided not just by a text or image prompt (for conditionally guided inference), but also by the modes that have been distilled in all of the preceding time steps. This allows ample time for the model to adjust and correct. However, for a white noise model, the conditioning of future mode-generation based on past mode generation is lopsided. Low-frequencies are unconstrained so they are largely free parameters determined by the initial noise injection and conditional guidance. High frequencies generate last so they are highly constrained by the low-frequency modes that have already been generated in previous time steps. In this sense, the model tends to treat the high-frequency details as ``fillers", over-constrained by the lower frequency modes that came before them. 

Combining the two effects, the white noise diffusion model generates high-frequency modes late in the inference process to fill in the blanks inside the large blocks of color and value that were generated in previous steps, and the model has a comparatively narrow window of timesteps where the high-frequency modes are generated. The net effect is an oversimplification of the high-frequency details, often resulting in the smoothed-out, airbrushed look that is characteristic of many diffusion-based image generators.

How can Cloud Diffusion improve the details of generated images? As we've argued, Cloud Diffusion models don't generate images sequentially from high to low frequency. At each step of the reverse diffusion process, noise is removed uniformly across all frequencies so that the signal-to-noise ratio is approximately independent of the wave-number in Fourier space. Consequently, high frequencies can be distilled at any time step. Moreover, continuing with the Bayesian interpretation where the predicted image generated cumulatively in all previous timesteps informs the prediction for the next time step, in a Cloud Diffusion Model, high frequency modes that have been generated in previous timesteps can now inform the predictions for low frequency modes in subsequent timesteps and vice-versa. This leads to more accurate correlations between the high and low frequency modes. The net result of the two effects is improved high-frequency fidelity.

\subsection{Improved Conditional Guidance}
Diffusion Models are typically guided by additional information, such as a text prompt or an external image, injected into the model throughout the reverse diffusion process \cite{dhariwal2021diffusion, saharia2022photorealistic, ho2022classifierfreediffusionguidance}. Here we argue that in some situations, the Cloud Diffusion architecture could lead to improvements in the conditional guidance over white noise diffusion models. 

The reasons for improved guidance again boils down to differences between the paths taken by each model from the noise distribution to the predicted image distribution. While classic white noise diffusion models build images from low to high frequency, Cloud Diffusion Models build images more holistically. How can this improve conditional guidance? Since white noise diffusion models generate high frequency modes at late time steps, prompts that condition primarily the high frequency modes can also only have impact at late time steps. By contrast, the same prompt in a Cloud Diffusion Models can condition the inference of all modes, at all time steps. 

\begin{figure}[htbp]
    \centering
    \includegraphics[width=0.9\linewidth]{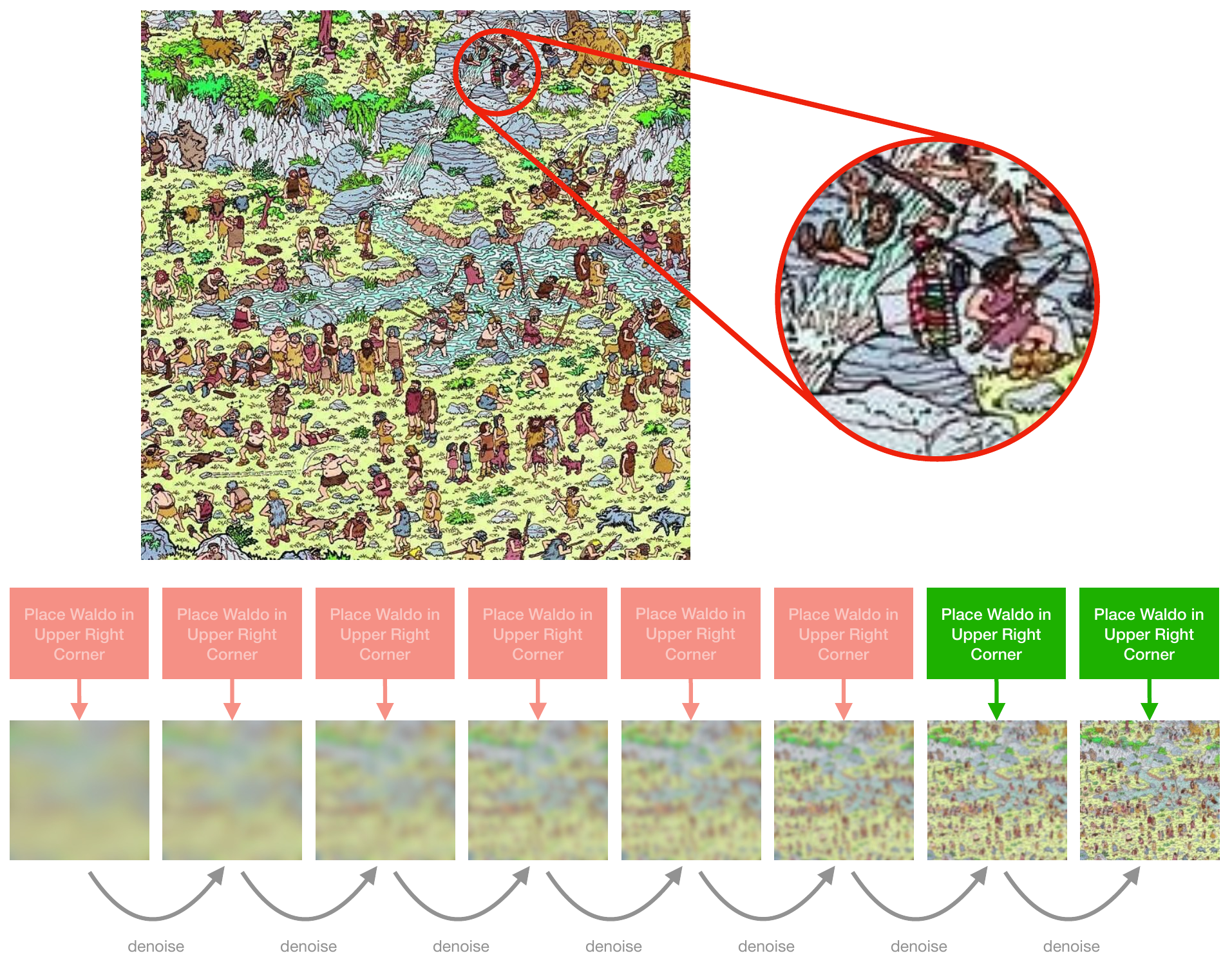}
    \caption{In this hypothetical scenario, a white-noise diffusion model generates a {\it Where's Waldo?} image via a text prompt. As Waldo is constructed primarily of high-frequency modes, the prompt is only effective at late timesteps. Consequently, the model can't adjust low-frequency modes conditioned on the generated Waldo image in previous timesteps.}
    \label{DenoisingWaldo}
\end{figure}

Let's flesh out the argument with an example. Suppose you are trying to train a model to produce images in the vein of {\it Where's Waldo?} \cite{Waldo} as shown in Figure \ref{DenoisingWaldo}. These images are unique in that most of the information content in the image is contained in the high-frequency modes. Indeed, Waldo himself is a high-frequency construct --- if you send a typical {\it Where's Waldo?} image through a low-pass filter, the location of Waldo himself will be completely lost. A well-constructed Waldo image will have visual distractions diverting the eye away from the actual location of Waldo, some of which may be composed of lower frequency modes than those comprising Waldo himself. Consequently, in a conditionally-guided model, the text-prompt can only condition, say, the location and details of the Waldo image at late time steps. The model cannot condition the lower frequency modes based on the actual generated Waldo image. It can only rely on the limited information in the text prompt. 
In general, the classic white noise diffusion models can be hampered in their ability to produce faithful images based on text and image prompts that conditionally guide the generation of high-frequency modes. By contrast, Cloud Diffusion Models may improve this type of high-frequency conditioning because high-frequency modes can be generated at any timestep in the reverse diffusion process. Consequently, the model can generate the high-frequency modes comprising Waldo early, which can then refine and serve to condition the generation of low-frequency modes, such as the overall composition and visual distractions, in subsequent timesteps.

\section{Concluding Remarks}
In this work we have outlined the architecture for a new type of diffusion model that uses the low-order statistical properties of the image set it seeks to emulate in the forward diffusion process. We've presented the theoretical foundation and detailed the motivation for the model, arguing that Cloud Noise can improve the quality, speed, and controllability of the diffusion paradigm. In a follow-up work, {\it``Cloud Diffusion Part 2: Training and Inference"}, we put the theory to practice, building a Cloud Diffusion Model, training it, and generating images. 

\section*{Acknowledgments}
I would like to thank Matt Patterson for early discussions on scale-invariant noise and its potential impact on diffusion models, Andrew Silberfarb for discussions of diffusion models and wider impact, and Sander Dieleman for discussions regarding his insightful blog post \cite{DielemanBlogPost} and sequential ordering the classic diffusion models follow from low-to-high frequency during inference. 

This paper was adapted from a previous work in a GitHub repository \cite{RandonoCloudDiffusionRepo} that was first publicly posted in 2024 and re-posted in a more polished form in April 2025. The excellent work of Falck, Karmalkar, and team at Microsoft \cite{falck2025fourier} that follows similar reasoning was recently brought to my attention, and I have been in correspondence with some of the authors. Both bodies of work were developed entirely independently, and there is no indication that either party was aware of the others' work while it was being developed. I take this as a sign that we are all on the right track!

\printbibliography
\end{document}